\documentclass[conference]{IEEEtran}
\IEEEoverridecommandlockouts
\usepackage{cite}
\usepackage{amsmath,amssymb,amsfonts}
\usepackage{graphicx}
\usepackage{textcomp}
\usepackage{xcolor}
\usepackage[bookmarks=false]{hyperref}
\usepackage{amsthm}
\usepackage{url}            
\usepackage{booktabs}       
\usepackage{amsfonts}       
\usepackage{nicefrac}       
\usepackage{microtype}      
\usepackage{color}
\usepackage{tabularx}
\usepackage{soul}
\usepackage{array}
\usepackage{amsmath}
\usepackage{longtable}
\usepackage{algorithm}
\usepackage{algpseudocode}
\usepackage{mathtools}
\usepackage{wrapfig}
\usepackage{multirow}
\usepackage{amsfonts}
\usepackage{subcaption}

\newtheorem{Assumption}{Assumption}
\newtheorem{Lemma}{Lemma}
\newtheorem{Proposition}{Proposition}
\newtheorem{Definition}{Definition}

\newcommand{\tabincell}[2]{\begin{tabular}{@{}#1@{}}#2\end{tabular}}  

\algnewcommand{\algorithmicforeach}{\textbf{for each}}
\algdef{SE}[FOR]{ForEach}{EndForEach}[1]
  {\algorithmicforeach\ #1\ \algorithmicdo}
  {\algorithmicend\ \algorithmicforeach}
  
\DeclarePairedDelimiter\abs{\lvert}{\rvert}

\def\BibTeX{{\rm B\kern-.05em{\sc i\kern-.025em b}\kern-.08em
    T\kern-.1667em\lower.7ex\hbox{E}\kern-.125emX}}

\begin{document}

\title{A Primal-Dual Subgradient Approach \\for Fair Meta Learning
}


\author{\IEEEauthorblockN{Chen Zhao, Feng Chen, Zhuoyi Wang, Latifur Khan}
\IEEEauthorblockA{\textit{Department of Computer Science} \\
\textit{The University of Texas at Dallas}\\
Richardson Texas, USA \\
\{chen.zhao, feng.chen, zhuoyi.wang1, lkhan\}@utdallas.edu}
}

\maketitle

\begin{abstract}
The problem of learning to generalize on unseen classes during the training step, also known as few-shot classification, has attracted considerable attention. Initialization based methods, such as the gradient-based model agnostic meta-learning (MAML) \cite{Finn-ICML-2017-(MAML)}, tackle the few-shot learning problem by ``learning to fine-tune”. The goal of these approaches is to learn proper model initialization, so that the classifiers for new classes can be learned from a few labeled examples with a small number of gradient update steps. Few shot meta-learning is well-known with its fast-adapted capability and accuracy generalization onto unseen tasks\cite{wang2020few}. Learning fairly with unbiased outcomes is another significant hallmark of human intelligence, which is rarely touched in few-shot meta-learning. 
In this work, we propose a \underline{P}rimal-\underline{D}ual \underline{F}air \underline{M}eta-learning framework, namely PDFM, which learns to train fair machine learning models using only a few examples based on data from related tasks. The key idea is to learn a good initialization of a fair model's primal and dual parameters so that it can adapt to a new fair learning task via a few gradient update steps. Instead of manually tuning the dual parameters as hyperparameters via a grid search, PDFM optimizes the initialization of the primal and dual parameters jointly for fair meta-learning via a subgradient primal-dual approach. 
We further instantiate an example of bias controlling using decision boundary covariance (DBC) \cite{Zafar-AISTATS-2017} as the fairness constraint for each task, and demonstrate the versatility of our proposed approach by applying it to classification on a variety of three real-world datasets. Our experiments show substantial improvements over the best prior work for this setting. 
\end{abstract}
\begin{IEEEkeywords}
dual subgradient, dual decomposition, meta-learning, fairness, few shot 
\end{IEEEkeywords}

\section{Introduction}
In contrast to the conventional machine learning systems, the ability to learn from a handful of examples is one of the critical characteristics of human intelligence. Learning quickly yet remains a daunting challenge for artificial intelligence, which receives significant attention from the machine learning community, especially when it needs to transfer knowledge from a given distribution of tasks onto unseen ones.
To address this challenge, meta-learning (\textit{a.k.a} learning to learn) leverages the transferable knowledge learned from previous tasks, then adapts to new environments rapidly with a few training examples. The goal of a few-shot meta-learning problem is to minimize generalization error across a distribution of tasks with few training examples (\textit{i.e.} few-shot). This technique has demonstrated success in both supervised learning, such as few-shot regression\cite{Finn-ICML-2017-(MAML),yoon2018}, classification\cite{Vinyals-NIPS-2016-(MatchingNet),Snell-NIPS-2017-(ProtoNet)}, and reinforcement learning\cite{xu2018meta} settings.

There are several lines of meta-learning algorithms for base learners, nearest neighbors based methods \cite{Vinyals-NIPS-2016-(MatchingNet), Snell-NIPS-2017-(ProtoNet)} which address the problem by ``learning to compare"; recurrent network-based methods \cite{Ravi-ICLR-2017} that instantiates the transferable knowledge as latent representations, and gradient-based methods \cite{Finn-ICML-2017-(MAML), Finn-NIPS-2018, Finn-ICML-2019, Nichol-arXiv-Reptile-2018, Rusu-ICLR-2019} that aim to learn proper model initialization for all tasks, such that the summation query errors is minimized and further the meta-parameter is adapted to novel tasks using a few optimization steps. Despite their early success in the few-shot application, to the best of our knowledge, most of the existing meta-learning algorithms ignore to mitigate the notion of fairness in tasks and thus lack the capability of fairness generalization on new tasks. 


Machine learning models trained to output prediction based on historical data will naturally inherit the past biases, with the biased input, the main goal of training an unbiased model is to make the output fair. In other words, the predictions are statistically independent of protected variables (\textit{e.g.} race and gender) \cite{Zliobaite-arXiv-2015}. Such models could be enhanced by masking some attributes to the decision-maker, however, as many attributes may be correlated with the protected one \cite{Zemel-ICML-2013}. Moreover, techniques in the area of fairness learning are incapable of adapting deep learning models on fairness to new tasks. This paper's motivation is: can we develop meta-learning methods that adapt deep learning models on both generalization accuracy and fairness to unseen tasks? 




\begin{figure}
    \centering
    \includegraphics[width = \linewidth]{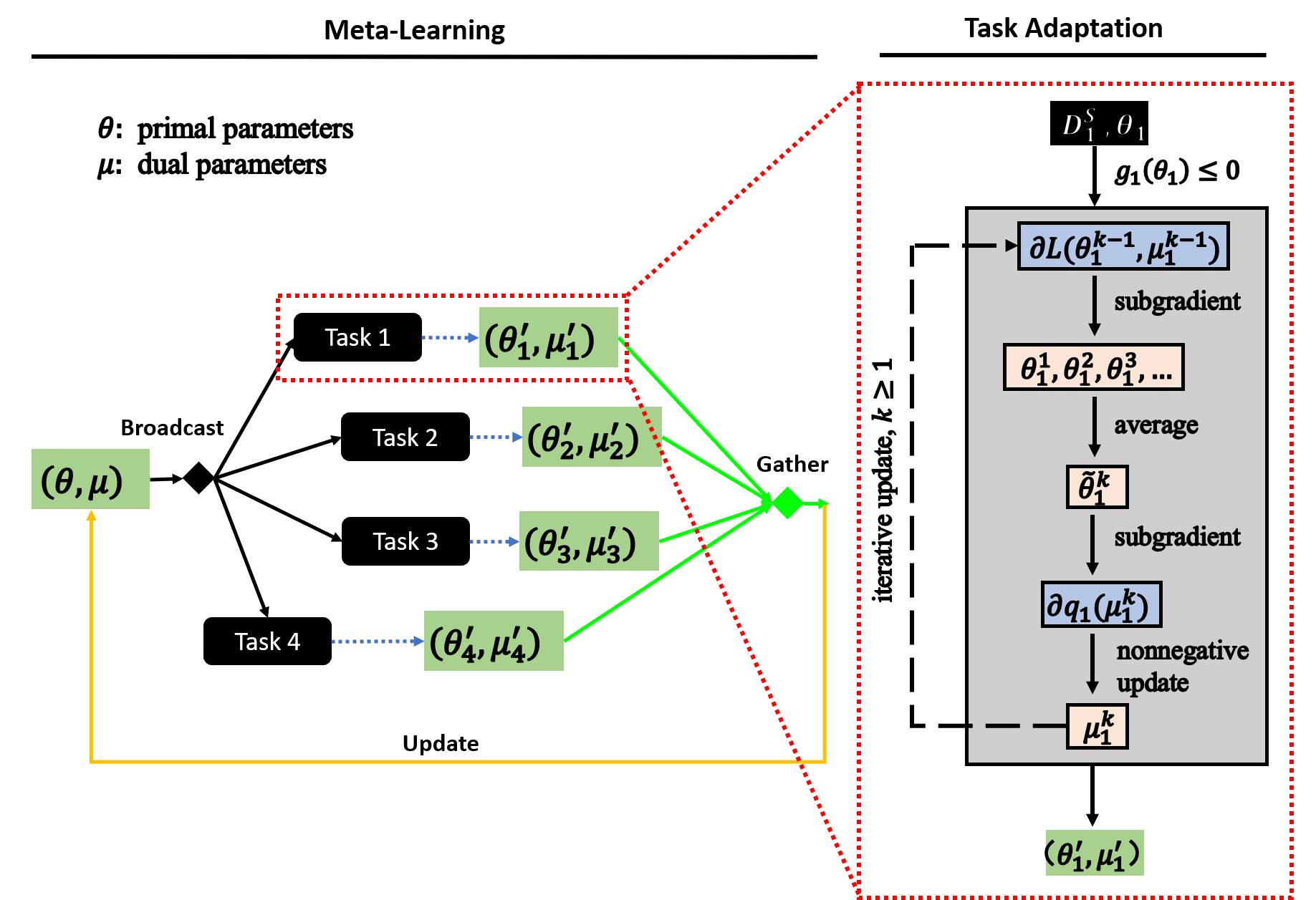}
    \vspace{-4mm}
    \caption{Schematic of our proposed PDFM pipeline. (Left) The global meta-parameters $(\theta, \mu)$ are sent to each task and each task optimizes in parallel to find a good task-specific primal-dual pair, \textit{e.g. }$(\theta'_1, \mu'_1)$, that is approximated by an averaging scheme dual subgradient algorithm presented on the right. Query losses and fairness are gathered and utilized to update the meta-initialization pair. (Right) A few-shot unfairness prevention approach is shown. In the meta-training stage, in each task, support loss is optimized under a fairness constraint which performs a trade-off between losses and fairness. The inner loop dual subgradient algorithm ensures that the duality gap of each task is minimum.}
    \label{fig:overview}
    \vspace{-5mm}
\end{figure}

This paper bridges areas of few-shot meta-learning and unfairness prevention and formulate this problem by enhancing the meta-learning model with fairness constraints. More concretely, for each task during the training stage, it is constrained with a task-specific fair inequality, which ensures the independent effect of the protected variable on task predictions. In the support set during the training process, the overall proportion of members in a protected group would receive predictions, which are identical to the proportion of the population as a whole. To this end, we resort to a dual subgradient algorithm with an averaging scheme for each task. It approximately optimizes a pair of task-specific primal and dual parameters, which minimizes the summation of query losses and fairness constraints are satisfied simultaneously. In contrast to the grid search technique, we consider Lagrange multipliers as dual variables that they are optimized to minimize the duality gap between the primal and dual functions.

Furthermore, instead of updating the meta-parameter from the outer loop (such as MAML \cite{Finn-ICML-2017-(MAML)}), in our work, inspired by the concept of resource allocation from economics, we propose a pair of primal-dual meta-parameters, which could be optimized iteratively through a dual decomposition \cite{Palomar-dual_decomposition-2007, Raffard-dual_decomposition-2004} and divided into \textit{broadcast} and \textit{gather} steps. We apply such decomposition to leverage the observation that problems can be decomposed into some sub-problems, and then introduce fairness constraints to enforce the notion of agreement between solutions to the different issues. The agreement constraints are incorporated using Lagrange multipliers, and an iterative algorithm is used to minimize the resulting dual. As shown in Figure \ref{fig:overview}, the interplay between the inner-algorithm (task-level) and the meta-algorithm plays a key role in our work. The former one is used to compute a good approximation of the meta-subgradient, and supplied to the latter. Finally, another key merit of this paper is that we derive an efficient and theoretically grounded analysis for the proposed meta-learning approach. Besides, we instantiate an example of decision boundary covariance (DBC) as the fairness constraint for justification, such constraint indicates the covariance between the protected variable and the signed distance from the feature vectors to the decision boundary \cite{Zafar-AISTATS-2017}. We demonstrate our proposed approach's versatility on a variety of three real-world datasets and extensive experiments to show substantial improvements over the best prior work. 

In summary, the main contributions of this paper is threefold:
\begin{itemize}
    \item We propose a novel \underline{P}rimal-\underline{D}ual \underline{F}air \underline{M}eta-learning framework, namely PDFM, in which a good pair of meta-parameters is approximately optimized. Our framework efficiently controls biases for each task, and ensures the generalization capability of both accuracy and fairness onto unseen tasks. 
    \item We further implement two optimized strategies for inner loop and meta-subgradient update. Specific and theoretically grounded analysis for the proposed strategies justifies the efficiency and effectiveness of them.
    \item Finally, we validate the performance of our approach with state-of-the-art techniques on three real-world datasets. Our results demonstrate the proposed approach is capable of mitigating biases, generalizing accuracy and fairness to unseen tasks with the minimized input training data.
\end{itemize}

\section{Related Work}
Meta-Learning based on few-shot studies that trained models to make it quickly adapt to new tasks, under a few labeled samples. Several recent approaches have made significant progress in meta-learning \cite{Bengio-ICLR-2020, Yao-ICML-2019, Tseng-ICLR-2020, Lian-ICLR-2020}. Previous algorithms majorly focus on the metric-based idea, which aim to learn an embedding space between query and support examples, where similar instances are closer and different ones are further apart\cite{Vinyals-NIPS-2016-(MatchingNet),Snell-NIPS-2017-(ProtoNet), sung2018learning, wang2019robust}. For example, the Matching-Net\cite{Vinyals-NIPS-2016-(MatchingNet)} employed ideas from k-nearest neighbors and metric learning based on a feature encoder to extract embedding in the context of the support set, and Prototypical networks \cite{Snell-NIPS-2017-(ProtoNet)} learn a metric space in which classification is able to be performed by computing Euclidean distances to prototype representations of each class.

In addition, gradient descent based algorithms \cite{Finn-ICML-2017-(MAML),Ravi-ICLR-2017,Finn-NIPS-2018,Rusu-ICLR-2019,Antoniou-ICLR-2019} aim to learn good model initialization so that the meta-loss is minimum.
They tend to meta-learn an initial set of weights for neural networks, and quickly adapted to new task with just a few steps of gradient descent, which could achieve good generalization over new tasks by encoding prior knowledge. Some existing work such as Franceschi et al.\cite{franceschi2018bilevel} also provide convergence guarantees for gradient-based meta-learning with strongly-convex functions. Despite methods in the area of meta-learning have been shown effective for adaption of deep learning models on generalization accuracy to new tasks, our experiments show such state-of-the-arts have difficulties in adaption on fairness.


Fairness researchers develop machine learning algorithms that would produce predictive models, ensuring that those models are free from biases. Standard predictive models, induced by machine learning and data mining algorithms, may discriminate groups of entities because (1) data bias comes from data being collected from different sources, or (2) dependence on sensitive attributes was identified in the data mining community \cite{Calders-ICDM-2013}. Based on the taxonomy by tasks, fairness learning can be typically categorized to classification\cite{Feldman-KDD-2015,Hardt-NIPS-2016,Zafar-AISTATS-2017}, regression\cite{Calders-ICDM-2013,Berk-FATML-2018,Zhao-ICDM-2019}, clustering\cite{Gondek-KDD-2005}, and recommendation\cite{Kamishima-RR-w-2017,singh2018} works. Even though techniques for unfairness prevention on classification were well developed, to the best of our knowledge, the majority of existing fairness-aware machine learning algorithms are under the assumption of giving abundant training examples. Learning quickly, however, is another significant hallmark of human intelligence.

Several recent approaches have been developed in fair meta-learning \cite{Slack-FAT-2019, Zhao-ICKG-1-2020, Zhao-ICKG-2-2020}. These methods focus on studies of fairness generalization onto unseen tasks by adding an uniformed fairness regularizer to each task. In addition, Lagrange multipliers were consider as hyperparameters and they were manually tuned by grid search. However, such prior studies suffer from limitations that (1) the trade-off parameter is valued the same for each task, and (2) hence there is a big room for improvement on the generalization of both accuracy and fairness onto new tasks. In this paper, to overcome such limitations, we develop a novel fair meta-learning framework. Each task is underwent a task-specific soft fairness constraint. Besides, we consider Lagrange multipliers as dual variables and hence, instead of grid search, they are optimized to minimize the duality gap between the primal and dual functions.



\section{Methodology}
\subsection{Problem Setting}
Let $\mathcal{Z=X \times Y}$ be the data space, where $\mathcal{X}\subset \mathbb{R}^n$ is the input space, $\mathcal{Y} = \{1,2,...,N\}$ means a sequence of discrete classes of the output space, and $N$ is the number of classes. Meta-learning for few-shot learning aims to train a meta-learner which is able to learn on a large number of various tasks from a small amount of data. Gradient based meta-learning frameworks, such as Model-Agnostic Meta-Learning (MAML) \cite{Finn-ICML-2017-(MAML)}, lead to state-of-the-art performance and fast adaptation to unseen tasks. More precisely, the goal of MAML is to estimate a good meta-parameter $\mathbf{\theta}\in\Theta$ such that the summation of empirical risks for each task is minimized. Throughout this work, the $\Theta$ will be a closed, convex, non-empty subset of an Euclidean space.

In this work, we consider a collection of supervised learning tasks $\mathcal{T} = \{(\mathcal{D}_t^{S}, \mathcal{D}_t^{Q})\}_{t=1}^T$ which distributions over $\mathcal{Z}$ and $T$ is denoted as the number of tasks. $\mathcal{T}$ is often referred to as a meta-training set as well as an episode $(\mathcal{D}_t^{S}, \mathcal{D}_t^{Q})$ explicitly contains a pair of a support (\textit{i.e.} $\mathcal{D}_t^{S}$) and a query (\textit{i.e.} $\mathcal{D}_t^{Q}$) data sets. For each task $t\in\{1,2,...,T\}$, we let $\{\mathbf{x}_{t,i}, y_{t,i}\}_{i=1}^m \in(\mathcal{X\times Y})$ be the corresponding task data, and $m$ is the number of datapoints in the support set. For example, standard few-shot learning benchmarks evaluate model in $N$-way $K$-shot classification tasks and thus $m=N\times K$ indicates, in the support set of the $t$-th task, it contains $N$ categories and each consists of $K$ datapoints. We emphasize that we need to sample without replacement, \textit{i.e.,} $\mathcal{D}_t^S \cap\mathcal{D}_t^Q=\emptyset$.

To study fairness generalization problem under meta-learning frameworks, a fairness constraint,  $g_t(\theta_t)\leq 0$, is considered in each task, where $t$ indicates task index. In researches of bias prevention, convexity of the constraint receives increasing attention in the machine learning fields \cite{Berk-FATML-2018, Zhang-AAAI-2019, Goel-AAAI-2018}. For this purpose, in this paper, we assume that convexity of task constraints always holds.

\subsection{Model-Agnostic Meta-Learning with constraints}
Meta-learning approaches for few-shot learning aim to minimize the generalization error across a distribution of tasks sampled from a task distribution. It is often assume that the support and query sets of a task are sampled from the same distribution. In our work, for each single task, the objective is to minimize the predictive error $\mathcal{L}^{inner}$ such that it is constrained by $g_t$:

\begin{align}
\label{inner_problem}
    \theta'_t = Alg(\mathcal{D}^S_t, \theta) = \arg\min_{\theta_t\in\Theta} \quad &f_t(\theta_t;\theta) := \mathcal{L}^{inner}(\mathcal{D}^S_t,\theta_t; \theta) \nonumber\\
    \text{subject to} \quad &g_t(\mathcal{D}^S_t,\theta_t) \leq 0 
\end{align}

where $\mathcal{L}^{inner}: \mathbb{R}^n \rightarrow \mathbb{R}$ is a loss function, such as cross-entropy loss for classification problems and $\theta_t$ is the model parameter at task $t$, which is initialized with $\theta$. $Alg(\mathcal{D},\theta)$ corresponds to one or multiple steps of gradient descent initialized at $\theta$.
$g: \mathbb{R}^n \rightarrow \mathbb{R}$ is an appropriate complexity function ensuring the existence and the uniqueness of the above minimizer. A point $\theta_t$ in the domain of the problem is feasible if it satisfies the constraint $g_t(\theta_t)\leq 0$. 

\begin{Assumption} (Task Loss and Constraint).
\label{assump1:task-loss-constraint}
Let $f_t(\theta_t)$ be a convex real-valued function for any $\theta_t\in\Theta$. Let $\Gamma(\Theta)$ be a set of proper, closed and convex function over $\Theta$ and $g_t\in\Gamma(\Theta)$ be such that, for any $\theta_t\in\Theta$, $g_t(\theta_t)$ is convex over $\mathbb{R}^n$, $\inf_{\theta_t\in\Theta} g_t(\theta_t)=0$ and, for any $\theta_t\notin \Theta$, \normalfont{dom}$(g_t(\theta_t))=\emptyset$.
\end{Assumption}

The optimal value of the Eq.(\ref{inner_problem}) is denoted as $f_t^*$, which is assume to be finite and is achieved at an optimal and feasible solution $\theta_t^*$, \textit{i.e.} $f_t^*=f_t(\theta_t^*)$. The goal of training a single task is to output local parameter $\theta_t$ given the meta-parameter $\theta$ such that it minimizes the task loss $f_t(\theta_t)$ subject to the task constraint $g_t(\theta_t)\leq 0$. Next, to update the meta-parameter, we minimize the generalization error $\mathcal{L}^{meta}$ using query sets across every tasks in the batch such that query constraints for all tasks are satisfied. Formally, the learning objective across all tasks is

\begin{align}
\label{meta_problem}
    &\min_{\theta\in\Theta} \quad \mathcal{L}^{meta} = \sum_{t=1}^T f_t(\theta'_t;\theta) := \sum_{t=1}^T \mathcal{L}^{inner}(\mathcal{D}^Q_t,Alg(\mathcal{D}^S_t, \theta)) \nonumber \\
    &\text{subject to} \quad \sum_{t=1}^T g_t(\mathcal{D}^Q_t, Alg(\mathcal{D}^S_t, \theta)) \leq 0 
\end{align}

where $\theta'_t=\arg\min_{\theta_t\in\Theta, g_t(\theta_t)\leq 0}f_t(\theta_t)$ is a local optimum of each task $t$. Here, for the purpose of optimization with simplicity, the constraint of Eq.(\ref{meta_problem}) is approximated, which originally takes the form of a sequence $g_t(\mathcal{D}^Q_t, Alg(\mathcal{D}^S_t, \theta))\leq 0$, where $t=1,...,T$. In this setting, the meta-objectives and the consequently their subgradients used by the meta-algorithm are dependent on the properties of the inner algorithm. We will show the algorithm details and analysis in the following sections.

\subsection{Primal and Dual Formulation}
Our approach aims to optimize a pair of meta-parameters (\textit{i.e.} primal and dual variables) as model initialization, instead of using the conventional grid search technique \cite{Slack-FAT-2019, Zhao-ICKG-1-2020, Zhao-ICKG-2-2020}. It consists of two nested primal-dual algorithms, one operating within each task and another across all tasks. In this section, we briefly recall from the primal-dual interpretation of the algorithm framework and such interpretation will be used in the subsequent analysis for both inner and meta problems.

To recover the primal optimal solution of Eq.(\ref{inner_problem}), we use the Lagrange duality theory to relax the primal problem by its constraints, and the Lagrangian function is 

\begin{align*}
    L(\theta_t, \mu_t) = f_t(\theta_t) + \mu_t^T g_t(\theta_t)
\end{align*}

where $\mu_t\in\mathbb{R}_+^m$ is the Lagrange multiplier (or dual variable). The dual function hence is defined as 

\begin{align*}
\label{inner_dual_function_q(mu)}
    q_t(\mu_t) = \inf_{\theta_t\in\Theta} L(\theta_t, \mu_t) = \inf_{\theta_t\in\Theta} \{f_t(\theta_t) + \mu_t^T g_t(\theta_t)\}
\end{align*}

Since the dual function $q_t(\mu_t)$ is a pointwise affine function of $\mu_t$, we thus can maximize the dual function to obtain a tightest lower bound of the optimal primal $f_t^*$ and through out this paper, we assume $f_t^*$ is finite. 
The goal is to obtain the dual optimal value $q_t^*$ at $\mu_t^*$, such that the duality gap, \textit{i.e.} $f_t^*-q_t^*$, is as small as possible. Zero duality gap thus indicates that the optimal values of the primal and dual problems are equal, \textit{i.e.} $f_t^*=q_t^*$. Due to space limit, the same idea is applied to solve Eq.(\ref{meta_problem}). The Lagrangian function of the outer loop is hence parameterized by the meta-pair $(\theta, \mu)$ and the goal is to find a good pair of initializations by optimizing a max-min problem.  


\subsection{Update Task-Specific Model-Parameters via Dual Subgradient}
In order to find a good pair of meta-parameters $(\theta, \mu) \in \Theta \times \mathbb{R}^m_+$, such that constraints of all tasks can be satisfied and generalization error is minimized. To this end, in this section, we provide an approximate solution to the inner task of Eq.(\ref{inner_problem}) by proposing a task-level dual subgradient algorithm. This method takes in the meta-parameter pair from the previous outer (or meta) loop and the task-specific (or local) primal and dual parameters are then iterative updated using the support data of the single task.

In the subsequent development, to solve the dual problem of Eq.(\ref{inner_problem}) for a single task, we consider a subgradient algorithm with a constant step size $\alpha\succ 0$ to update the dual solution iteratively:

\begin{align}
    \mu_t^{k} = [\mu_t^{k-1} + \alpha^T g_k]^+
\end{align}

where $[u]^+$ denotes the  projection of $[u]$ on the nonnegative orthant in $\mathbb{R}^m_+$, namely $[u]^+ = (\max\{0,u_1\}),...,\max\{0,u_m\})$, $k=1,2,...$ is the index of iterations, subscript $t$ is the task index number, and $\mu_t^0\succ 0$ is an initial dual point. The subgradient iterate $g_k$ is a subgradient of the dual function $q_t$ at a given $\mu_t^k\succeq 0$:

\begin{align}
    g_k = g_t(\Tilde{\theta}_t^k) \in \partial q_t(\mu_t^k) = \text{conv}(\{g_t(\Tilde{\theta}_t^k)| \Tilde{\theta}_t^k\in\Theta_{\mu_t^k}\}) 
\end{align}

where $\Theta_{\mu_t^k} = \{\Tilde{\theta}_t^k\in\Theta|q_t(\mu_t^k)=f_t(\Tilde{\theta}_t^k)+(\mu_t^k)^T g_t(\Tilde{\theta}_t^k)\}$ and conv$(Y)$ denotes the convex hull of a set $Y$. Although a general dual subgradient method can generate near-optimal dual solutions with a sufficiently small step size and a large number of iterations, it does not directly provide primal solutions which are of our interest. But even worse, it may fail to produce any useful information. Motivated by this reason, we apply an averaging scheme to the primal sequence $\{\theta_t^k\}$ to approximate primal optimal solutions. In particular, the sequence $\{\Tilde{\theta}_t^k\}$ is defined as the averages of the previous vectors through $\theta_t^0$ to $\theta_t^{k-1}$, 

\begin{align}
\label{eq:average-of-the-primal}
    \Tilde{\theta}_t^k = \frac{1}{k}\sum_{i=1}^{k-1}\theta_t^i, \quad \forall k\geq 1
\end{align}

where the corresponding primal feasible iterate $\theta^k$ is given by any solution of the set. 

\begin{align}
    \theta_t^k \in \arg\min_{\theta_t\in\Theta}\{f_t(\theta_t^{k-1})+(\mu_t^{k-1})^T g_t(\theta_t^{k-1})\}
\end{align}

As the subgradient method can usually generate a reasonable estimation of the dual optimal solutions within several iterations, approximate primal solutions are obtained accordingly. The constant stepsize $\alpha$ is a simple hyperparameter for controlling, then through choosing an appropriate value of $\alpha$, the proposed Algorithm \ref{alg:dual_subgradient} is able to approach the optimal value arbitrarily close within a small finite number of steps.

\begin{algorithm}[!t]
\caption{Update Model-parameters of Task $t$ using Dual Subgradient Method}
\label{alg:dual_subgradient}
\textbf{Require}: $\theta\in\Theta, \mu\in\mathbb{R}^m_+$: prime and dual initializations\\
\textbf{Require}: $\alpha\succ 0, \gamma\succ 0$: learning rate \\
\textbf{Require}: $q>0$: a small number of subgradient update steps

\begin{algorithmic}[1]
\State $\mu_t^0 \leftarrow \mu$, $\theta_t^0 \leftarrow \theta$
\State Initialize an empty array $a=\emptyset$
\For{$k=1,2,...$}
    \For{$q=1,2,...$}
        \State Evaluate the primal feasible subgradient $\Bar{\nabla} \in \nabla_{\theta_t^{k-1}}\{f_t(\theta_t^{k-1})+(\mu_t^{k-1})^T g_t(\theta_t^{k-1})\}$
        \State $\theta_t^k \leftarrow \theta_t^{k-1} - \gamma^T\Bar{\nabla}$
    \EndFor
    \State Add $\theta_t^k$ in $a$ 
    \State Evaluate $\Tilde{\theta}_t^k$ by taking the average of previous vectors in $a$: $\Tilde{\theta}_t^k = \frac{1}{k}\sum_{i=0}^{k-1}\theta_t^i$
    \State Calculate the subgradient iterate $g_k = g_t(\Tilde{\theta}_t^k)$
    \State Update the dual solution $\mu_t^{k} = [\mu_t^{k-1} + \alpha^T g_k]^+$
\EndFor
\State \textbf{return} $(\theta'_t,\mu'_t)$, where $\theta'_t = \theta_t^k, \mu'_t = \mu_t^k$
\end{algorithmic}
\end{algorithm}
\setlength{\textfloatsep}{0pt}

Moreover, the dual subgradient schemes can be applied efficiently to approximate a solution to Eq.(\ref{inner_problem}). Specifically, it returns a good pair of task-level primal and dual parameters $(\theta'_t, \mu'_t)$. In the following section, due to the decomposable structure of the meta-learning framework for few-shot learning, meta-parameters $(\theta, \mu)$ are updated by minimizing the summation of query losses across all training tasks.

\subsection{Update Meta-parameters via Dual Decomposition}

In this work, inspired by the concept of resource allocation from economics \cite{Palomar-dual_decomposition-2007, Raffard-dual_decomposition-2004}, our model's goal is to estimate a good pair of primal-dual weight initialization $(\theta, \mu)$, such that both the meta-loss across tasks is minimum and constraints of all tasks are also satisfied. To this end, we update the pair of primal-dual initialization iteratively using a dual decomposition method that is normally considered as a special case of Lagrangian relaxation \cite{Rush-AIR-2012}. This method is typically simple and efficient, which can be divided into two steps for each iterate, \textit{i.e.} \textit{broadcast} and \textit{gather}. In the \textit{broadcast} step, the meta-dual parameter $\mu$ is sent to each of tasks $\mathcal{T}_t$. Through Algorithm \ref{alg:dual_subgradient}, local primal, and dual parameters $\theta_t$ and $\mu_t$ of a single task are iteratively optimized using few-shot support data. Query loss $f_t(\mathcal{D}_t^Q,\theta'_t)$ and fairness estimate $g_t(\mathcal{D}_t^Q,\theta'_t)$, therefore, are evaluated using query data set. In the \textit{gather} step, both query losses and fairness estimates collected across all tasks are applied to update primal and dual meta-parameters,

\begin{align}
\label{eq:dual-decomposition}
    \theta^{s+1} &\in \arg\min_{\theta\in\Theta} \sum_{t=1}^T f_t(\theta'_t;\theta^s) + \mu^s \sum_{t=1}^T g_t(\theta'_t;\theta^s) \\
    \mu^{s+1} &= [\mu^s + \beta\sum_{t=1}^T g_t(\theta'_t) ]^+
\end{align}

where $s=1,2,...$ is the index of the outer iteration and $\beta\succ 0$ is the stepsize. The full algorithm of the proposed approach is outlined in Algorithm \ref{alg:duality_maml}.

\section{Analysis}


\begin{algorithm}[!t]
\caption{The Primal-Dual Fair Meta-learning (PDFM) Algorithm}
\label{alg:duality_maml}
\textbf{Require}: $p(\mathcal{T})$: distribution over tasks\\
\textbf{Require}: $\eta\succ 0, \beta\succ 0$: learning rate
\begin{algorithmic}[1]
\State randomly initialize primal and dual meta-parameter, \textit{i.e.} $\theta\in\Theta$ and $\mu\in\mathbb{R}_+^m$ 
\While{not done}
    \State sample batch of tasks $\mathcal{T}_t\sim p(\mathcal{T}), t=1,2,...,T$
    \For{all $\mathcal{T}_t = \{\mathcal{D}^\mathcal{S}_t, \mathcal{D}^\mathcal{Q}_t\}$}
        \State Sample datapoints $\mathcal{D}_t^S=\{\mathbf{x}_t, \mathbf{y}_t\}$ from $\mathcal{T}_t$
        \State Compute adapted primal-dual parameters $\theta'_t$ and $\mu'_t$ using $\mathcal{D}_t^S$ by applying \textit{\textbf{Algorithm \ref{alg:dual_subgradient}}}
        \State Sample datapoints $\mathcal{D}_t^Q=\{\mathbf{x}_t, \mathbf{y}_t\}$ from $\mathcal{T}_t$ for the meta-update, where $\mathcal{D}_t^S \cap\mathcal{D}_t^Q=\emptyset$
        \State Evaluate query loss $f_t(\theta'_t)$ and query constraint $g_t(\theta'_t)$ using $\mathcal{D}^\mathcal{Q}_t$
    \EndFor
    \State Update $\theta$ and $\mu$ using Eq.(\ref{eq:dual-decomposition}).  \algorithmiccomment{Update Meta-parameters.}
\EndWhile
\end{algorithmic}
\end{algorithm}

Recall that the proposed averaging scheme used to approximate the task-specific primal-dual parameter pair is built upon the dual subgradient method with a constant stepsize. We denote the dual feasible set as $M=\{\mu_t|\mu_t\succeq 0, -\infty < q_t(\mu_t) < \infty \}$, and for every fixed $\mu_t\in M$, we have the solution set $\mathcal{C}\subset \Theta$ for $q_t(\mu_t)$.

\begin{Assumption} (Slater Condition and Bounded Subgradients)
\label{assump2}
The convex set $\Theta$ is compact (\textit{i.e.} closed and bounded). There exists a Slater point $\Bar{\theta}_t\in\Theta$, such that $g_j(\Bar{\theta}_t)<0, \forall j=1,2,...,m$, and exists $L>0, L\in\mathbb{R}$, such that $||g_k||<L, \forall k\geq 0$.
\end{Assumption}

When $f_t^*$ is finite, the Slater condition is sufficient for the existence of a dual optimal solution, and therefore the proposed task adaptation approach efficiently reduces the amount of feasibility violation at the approximate primal solutions. Furthermore, intuitively, 
bounded subgradients in \textit{Assumption} \ref{assump2} is satisfied when $L=\max_{\Tilde{\theta}_t\in\Theta}||g_t(\Tilde{\theta}_t)||$. 


\begin{Lemma}
\label{lemma 1}
If Assumption \ref{assump1:task-loss-constraint} and the continuity of $f_t(\theta_t)$ and $g_t(\theta_t)$ hold, there exists at least one optimal solution $\theta_\mu\in\mathcal{C}$. Furthermore, $\theta_\mu$ is unique if $f_t(\theta_t)$ is strictly convex, otherwise there may be  multiple solutions.
\end{Lemma}




Due to space limit, Lemma \ref{lemma 1} is easily proved using the \textit{Weierstrass Theorem} proposed in \cite{Bertsekas-book}. 
Next, for the averaged primal sequence $\{\Tilde{\theta}_t^k\}$, we show that it always converges when $\Theta$ is compact \cite{averagedDS-SJO-2009}.

\begin{Proposition}
Under Assumption \ref{assump2}, when the convex set $\Theta$ is compact, let the approximate primal sequence $\{\Tilde{\theta}_t^k\}$ be the running averages of the primal iterates given in Eq.(\ref{eq:average-of-the-primal}). Then $\{\Tilde{\theta}_t^k\}$ can converge to its limit $\Tilde{\theta}_t^*$.
\end{Proposition}

\begin{IEEEproof}
For simplicity, the subscript $t$ is hidden. 
To prove the convergence, we first show that $\{\Tilde{\theta}^k\}$ is a Cauchy sequence, \textit{i.e.} $\forall \epsilon>0$, there is a $K\in\mathbb{N}$ such that $||\Tilde{\theta}^{k'}-\Tilde{\theta}^k||<\epsilon, \forall k',k\geq K$. Given Eq.(\ref{eq:average-of-the-primal}), we can derive $\Tilde{\theta}^{k+1} = \frac{k}{k+1}\Tilde{\theta}^k + \frac{1}{k+1}\theta^k$. And hence $\Tilde{\theta}^{k+1} - \Tilde{\theta}^k = \frac{\theta^k-\Tilde{\theta}^k}{k+1}$. Since $\Theta$ is a compact convex set and we assume $k'>k$, we have $\theta^k, \Tilde{\theta}^k \in\Theta$ and $||\theta^k||, ||\Tilde{\theta}^k|| \leq M$, where $M\geq 0$. Iteratively, we have
\begin{align*}
    ||\Tilde{\theta}^{k'} - \Tilde{\theta}^k|| &= ||\Tilde{\theta}^{k'} - \Tilde{\theta}^{k'-1} + \cdots + \Tilde{\theta}^{k+1} - \Tilde{\theta}^k ||\\
    &= ||\frac{\theta^{k'-1} - \Tilde{\theta}^{k'-1}}{k'} + \cdots +  \frac{\theta^k-\Tilde{\theta}^k}{k+1}|| \\
    &\leq \frac{||\theta^{k'-1}||+|| \Tilde{\theta}^{k'-1}||}{k'} + \cdots + \frac{||\theta^k|| + ||\Tilde{\theta}^k||}{k+1} \\
    &\leq \frac{2M(k'-k)}{k+1}
\end{align*}
Therefore, for any arbitrary $\epsilon>0$, we let $\frac{2M(k'-k)}{k+1}<\epsilon$ and we have $||\Tilde{\theta}^{k'} - \Tilde{\theta}^k||<\epsilon, \forall k',k\geq K$. Thus, $\{\Tilde{\theta}^k\}$ is a Cauchy sequence. Furthermore, since a Cauchy sequence is bounded, there is a subsequence $b_n$ converging to the limit $L$ of it. For any $\epsilon>0$, there exists $n,m\geq K$ satisfying $||\Tilde{\theta}^n - \Tilde{\theta}^m|| < \frac{\epsilon}{2}$. Thus, there is a $b_k=\Tilde{\theta}^{m_k}$, such that $m_k\geq K$ and $||b_{m_k}-L||<\frac{\epsilon}{2}$.
\begin{align*}
    ||\Tilde{\theta}^n - L|| &= ||\Tilde{\theta}^n - b_k + b_k - L|| \\
                             &\leq ||\Tilde{\theta}^n - b_k|| + ||b_k - L|| \\
                             &< ||\Tilde{\theta}^n - \Tilde{\theta}^m|| + \frac{\epsilon}{2} < \epsilon
\end{align*}
Since $\epsilon$ is arbitrarily small, we proof that the sequence $\{\Tilde{\theta}^k\}$ converges to its limit $L = \Tilde{\theta}^*$ asymptotically.
\end{IEEEproof}

Besides, since the proposed Algorithm \ref{alg:duality_maml} is considered as an extended and modified version of  \cite{Finn-ICML-2017-(MAML)}, convergence of Algorithm \ref{alg:duality_maml} is guaranteed and detailed analysis is stated in \cite{Fallah-AISTATS-2020}. Accessing to sufficient samples, the running time of the proposed approach is $O(s\cdot k\cdot q)$ , where $s, k$ are respectively the number of outer and inner iterations, and $q$ is gradient steps of inner loop. For a $N$-way-$K$-shot learning, the best accuracy is achieved when $||\nabla \theta||\leq O(\Tilde{\sigma}/\sqrt{NK})$, where $\theta = \mathbb{E}_{\mathcal{T}\sim p(\mathcal{T})} l_{\mathcal{T}}(f_\theta)$, $l_\mathcal{T}$ is the query loss of task $\mathcal{T}$, $\sigma$ is a bound on the standard deviation of $\nabla L_t(\theta_t, \mu_t)$ from its mean $\nabla L(\theta, \mu)$, and $\Tilde{\sigma}$ is a bound on the standard deviation of estimating $\nabla L_t(\theta_t, \mu_t)$ using a single data point.

\section{A Classification Example in Unfairness Prevention}

In the previous section, we derived a theoretically principled algorithm under the assumption that the convexity always holds for both $f_t(\cdot)$ and $g_t(\cdot)$. However, many problems of interest in machine learning and deep learning have a non-convex landscape due to the non-linearity of neural networks, where theoretical analysis is challenging. Nevertheless, algorithms originally developed for convex optimization problems like gradient descent have shown promising results in practical non-convex settings. Taking inspiration from these successes, in this section, we respectively describe practical instantiations of our unfairness prevention for classification problems, and empirically evaluate the performance in Section VII.

Intuitively, an attribute affects the target variable if one depends on the other. Strong dependency indicates strong effects. Currently, most fairness criteria used for evaluating and designing machine learning models focus on the relationships between the protected attribute and the system output. For simplicity, we consider one binary protected attribute (\textit{e.g.} white and black) in this work. However, our ideas can be easily extended to many protected attributes with multiple levels. We thus modify the introduced setting by letting $\mathcal{Z=X\times Y}$ be the data space, where $\mathcal{X} = \mathcal{E} \cup\mathcal{S}$. Here $\mathcal{E} \subset \mathbb{R}^n$ is an input space, $\mathcal{S} = \{0,1\}$ is a protected space, and $\mathcal{Y} = \{0,1\}$ is an output space for binary classification. For each task $t\in\{1,2,...,T\}$, we let $\{\mathbf{e}_{t,i}, y_{t,i}, s_{t,i}\}_{i=1}^m \in(\mathcal{E\times Y\times S})$ be the corresponding task data and $m$ is the number of datapoints in the support set. In a $N$-way-$K$-shot classification problem, since we assume all the tasks to be binary labeled, in this example, all of our tasks are 2-way (\textit{i.e.} $N=2$). In referencing K-shot fairness, we mean that we are using $K$ training examples irrespective of class label, with the assumption that all tasks are 2-way. A fine-grained measurement to ensure fairness in class label prediction is to design fair classifiers by controlling the decision boundary covariance (DBC) \cite{Zafar-AISTATS-2017}.

\begin{Definition}[Decision Boundary Covariance \cite{Zafar-AISTATS-2017}]
The covariance between the protected variables $\mathbf{s}=\{s_i\}_{i=1}^h$ and the signed distance from the feature vectors to the decision boundary, $d_\mathbf{\theta}(\mathbf{e}) = \{d_\mathbf{\theta}(\mathbf{e}_i)\}_{i=1}^h$, 
\begin{align}
\label{dbc definition}
    DBC(\mathbf{s}, d_\mathbf{\theta}(\mathbf{e})) &= \mathbb{E}[(\mathbf{s-\Bar{s}})d_\mathbf{\theta}(\mathbf{e})] - \mathbb{E}[\mathbf{s-\Bar{s}}]\Bar{d}_\mathbf{\theta}(\mathbf{e}) \nonumber\\
    &\approx \frac{1}{h}\sum_{i=1}^h (\mathbf{s}_i-\mathbf{\Bar{s}})d_\mathbf{\theta}(\mathbf{e})
\end{align}
\end{Definition}

where $\mathbb{E}[\mathbf{s-\Bar{s}}]\Bar{d}_\mathbf{\theta}(\mathbf{e})$ is cancels out since $\mathbb{E}[\mathbf{s-\Bar{s}}]=0$ and $h=N\times K$ is the sample size of a support set of a single task. In a linear model for classification, such as logistic regression, the decision boundary is simply the hyperplane defined by $\theta^T\mathbf{e}=0$. A point $\theta_t$ in the domain of a task is feasible if it satisfies the constraint $g_t(\theta_t)\leq 0$. More concretely, $g_t(\theta_t)$ is defined by the definition of DBC in Eq.(\ref{dbc definition}), \textit{i.e.}
\begin{align}
    g_t(\theta_t) = \abs*{\frac{1}{2K}\sum_{\mathbf{s}_i, \mathbf{e}_i\sim\mathcal{T}_t}(\mathbf{s}_i-\Bar{\mathbf{s}})d_{\theta_t}(\mathbf{e}_i)}-c
\end{align}
where $c$ is a small positive fairness relaxation. To formalize the supervised classification problem in the context of meta-learning definitions, a cross-entropy loss function is used to describe the adapted loss over a support set for each task. Integrated with DBC fairness constraint, the classification problem of a single task is formulated as follow

\begin{align}
    \min_{\theta_t\in\Theta} \quad &f_t(\theta_t) = 
    \sum_{(\mathbf{e}^i, y^i)\sim\mathcal{T}_t} y^i\log \hat{y}(\mathbf{e}^i,\theta_t) \\
    &+(1-y^i)\log(1-\hat{y}(\mathbf{e}^i,\theta_t)) \nonumber\\
    \text{subject to} \quad &\abs*{\frac{1}{2K}\sum_{\mathbf{s}_i, \mathbf{e}_i\sim\mathcal{T}_t}(\mathbf{s}_i-\Bar{\mathbf{s}})d_{\theta_t}(\mathbf{e}_i)} \leq c \nonumber
\end{align}

where $(\mathbf{e}^i, y^i)$ are an input/output pair sampled from task $\mathcal{T}_t$ and $\hat{y}$ is a predicted outcome. The goal of a single task optimization is to approximate a good parameter pair $(\theta'_t, \mu'_t)$ by applying the proposed dual subgradient method and further pass the pair to evaluate accuracy and fairness (\textit{i.e.} DBC) over the query data. As the original meta-learning problem in Eq.(\ref{meta_problem}) is decomposed into a batch of single tasks, meta-parameters $(\theta, \mu)$ are iteratively updated using the proposed dual decomposition approach outlined in Algorithm \ref{alg:duality_maml}.

\section{Experimental Settings}
To validate our approach of unfairness prevention in few-shot meta-learning models, we conduct experiments
with three real-world datasets which are available from the UCI ML-repository.

\subsection{Data}

The \textbf{Adult} income dataset\cite{AdultDataSet-UCI-1994} contains a total of 34 tasks according to different countries and regions, totally 48,842 instances with 14 features (e.g., age, educational level) and a binary label, which indicates whether a subject’s incomes is above or below 50K dollars. We consider gender, \textit{i.e.} male and female, as the protected attribute.

\textbf{Communities and Crime} dataset \cite{CommunitiesandCrimeDataSet-UCI-1994} includes information relevant to crime (e.g., police per population, income) as well as demographic information (such as race and sex) in different communities across the U.S. We convert this dataset to a few-shot fairness setting by using each state as a different task. Following the same setting in \cite{Slack-FAT-2019}, since the violent crime rate is a continuous value, we convert it into a binary label based on whether the community is in the top 50\%  violent crime rate within a state. Additionally, we add a binary sensitive column that receives a protected label if African-Americans are the highest or second highest population in a community in terms of percentage racial makeup. 

\textbf{Bank Marketing} dataset \cite{Moro-bank-marketing-dataset-2014} contains a total 41,188 subjects, each with 20 attributes (\textit{e.g.} loan, housing, \textit{etc.}) and a binary label, which indicates whether the client has subscribed or not to a term deposit. In this case, we consider the marital status as the binary protected attribute, which is discretized to indicate whether the client is married or not. Since the dataset contains information of different months (\textit{i.e.} January to December) and dates (\textit{i.e.} Monday to Friday), we combine them as task labels and thus the dataset contains 50 tasks.

\begin{table}[!t]
\footnotesize
    \centering
    \captionof{table}{Key characteristics and statistics of real dataset. }
    \label{tab:key}
    \setlength\tabcolsep{3pt}
    \begin{tabular}{|c|c|c|c|}
        \hline
        Data & Adult & \tabincell{c}{Communities \\and Crime} & Bank\\ 
        \hline
        $s$ & $\{$M, F$\}$ & \tabincell{c}{$\{$Black,\\ non-Black$\}$} & \tabincell{c}{$\{$Married, \\non-Married$\}$}\\
        \hline
        $y$ & \tabincell{c}{income\\ \{$\geq \text{or} <50K$\}} & \tabincell{c}{crime rate\\ \{$\geq \text{or} <50\%$\}} & \tabincell{c}{deposit\\ \{Yes, No\}} \\
        \hline
        $\#$ of instance & 48,842 & 2,216 & 41,188 \\
        \hline
        tasks & countries & states & months and dates \\
        \hline
        $\#$ of total tasks & 34 & 46 & 50 \\
        \hline
        $\#$ of input features & 12 & 98 & 17\\
        \hline
        tasks for training & 22 & 30 & 40\\
        \hline
        tasks for validation & 6 & 8 & 5\\
        \hline
        tasks for testing & 6 & 8 & 5 \\
        \hline
        DBC & 0.043 & 0.052 & 0.067 \\
        \hline
        Discrimination & 0.195 & 0.214 & 0.028\\
        \hline
        Consistency & 0.485 & 0.222 & 0.377 \\
        \hline
    \end{tabular}
\end{table}

\subsection{Evaluation Metrics}
To evaluate the proposed techniques for fairness learning, we introduced two classic evaluation metrics to measure data biases. These measurements came into play that allows quantifying the extent of bias taking into account the protected attribute and were designed for indicating indirect discrimination.

\textbf{Discrimination} measures the bias with respect to the protected attribute $S$ in the classification:
\begin{align*}
    \text{Disc} =  \abs*{\frac{\sum_{i:s_i=1} \hat{y}_i}{\sum_{i:s_i=1} 1} - \frac{\sum_{i:s_i=0} \hat{y}_i}{\sum_{i:s_i=0} 1}}
\end{align*}
This is a form of statistical parity that is applied to the binary classification decisions. It measures the difference in the proportion of positive classifications of individuals in the protected and unprotected groups. $Disc=0$ indicates there is no discrimination.
    
\textbf{Consistency}\cite{Zemel-ICML-2013} compares a model's classification prediction of a given data item to its $k$-nearest neighbors:
\begin{align*}
    \text{Cons} = 1- \frac{1}{|D|k} \sum_{i=1}^{|D|}\abs*{\hat{y}_i - \sum_{j\in kNN (\mathbf{e}_i)} \hat{y}_j }
\end{align*}
where $|D|$ is the sample size, $k$ is the number of nearest neighbors, and a nearest neighbor is defined based on a similarity measure (\textit{i.e.} euclidean distance) of unprotected attributes $\mathbf{e}$. As demonstrated in \cite{Zemel-ICML-2013}, we applied the kNN function to the full set of examples to obtain the most accurate estimate of each point’s nearest neighbors. The consistency is a real number with a value of one signifying a fair prediction.

\subsection{Baseline Methods}
We evaluate all datasets – the proposed approach against various baselines – by comparing the results of generalization on both classification accuracy and fairness applied to:

\begin{table*}[!htbp]
\footnotesize
    \centering
    \captionof{table}{Consolidated overall result for few-shot classification.}
    \begin{tabular}{c|l|cccc|cccc|cccc}
        \hline
         & & \multicolumn{4}{c|}{\textbf{Adult Data}}  & \multicolumn{4}{c|}{\textbf{Communities and Crime}} & \multicolumn{4}{c}{\textbf{Bank Marketing}}\\
        \textbf{K} & \textbf{Approach} & \textbf{Acc} & \textbf{DBC} & \textbf{Disc} & \textbf{Cons}& \textbf{Acc} & \textbf{DBC} & \textbf{Disc} & \textbf{Cons}& \textbf{Acc} & \textbf{DBC} & \textbf{Disc} & \textbf{Cons}\\
        \hline
        - & Data & - & 0.043 & 0.195 & 0.485 & - & 0.052 & 0.214 & 0.222 & - & 0.067 & 0.028 & 0.377 \\
        \hline
        
        \parbox[t]{2mm}{\multirow{8}{*}{\rotatebox[origin=c]{90}{\textbf{5-shot}}}} 
         & MAML\cite{Finn-ICML-2017-(MAML)} & 82.1\% & 0.046 & 0.227 & 0.883 & \textbf{98.4\%} & 0.039 & 0.450 & 0.726 & 61.1\% & 0.026 & 0.122 & 0.884\\
         & Masked MAML & 79.9\% & - & 0.157 & 0.916 & 85.8\% & - & 0.322 & 0.846 & 57.6\% & - & 0.083 & 0.926 \\
         & pretrain & 76.5\% & 0.024 & 0.239 & 0.907 & 84.5\% & 0.030 & 0.337 & 0.815 & 57.1\% & 0.018 & 0.106 & 0.923\\
         & fair-MAML\cite{Zhao-ICKG-2-2020} & 59.7\% & 0.028 & 0.146 & 0.909 & 77.2\% & 0.026 & 0.358 & 0.758 & 56.2\% & 0.012 & 0.057 & \textbf{0.952}\\
         & F-MAML$_{dp}$\cite{Slack-FAT-2019} & \textbf{82.8\%} & 0.030 & 0.159 & 0.913 & 95.1\% & 0.039 & 0.442 & 0.757 & 59.3\% & 0.017 & 0.081 & 0.929\\
         & F-MAML$_{eop}$\cite{Slack-FAT-2019} & 79.5\% & 0.029 & 0.153 & 0.916 & 95.0\% & 0.041 & 0.387 & 0.775 & 57.0\% & 0.017 & 0.083 & 0.927 \\
         & LAFTR\cite{Madras-ICML-2018} & 72.0\% & 0.035 & 0.188 & 0.891 & 89.2\% & 0.050 & 0.440 & 0.787 & \textbf{62.1\%} & 0.030 & 0.100 & 0.865 \\
         & Ours & 78.2\% & \textbf{0.003} & \textbf{0.026} & \textbf{0.937} & 79.0\% & \textbf{0.013} & \textbf{0.200} & \textbf{0.893} & 55.9\% & \textbf{0.005} & \textbf{0.026} & 0.950 \\
        \hline
        
        \parbox[t]{2mm}{\multirow{8}{*}{\rotatebox[origin=c]{90}{\textbf{10-shot}}}} 
         & MAML\cite{Finn-ICML-2017-(MAML)} & 81.9\% & 0.045 & 0.211 & 0.900 & \textbf{99.6\%} & 0.038 & 0.463 & 0.760 & 59.7\% & 0.020 & 0.089 & 0.898\\
         & Masked MAML & 80.0\% & - & 0.143 & 0.930 & 86.5\% & - & 0.275 & 0.864 & 57.7\% & - & 0.059 & 0.941 \\
         & pretrain & 78.8\% & 0.023 & 0.125 & 0.923 & 83.2\% & 0.030 & 0.293 & 0.849 & 58.3\% & \textbf{0.008} & 0.039 & 0.969\\
         & fair-MAML\cite{Zhao-ICKG-2-2020} & 70.0\% & 0.030 & 0.146 & 0.925 & 83.6\% & 0.035 & 0.356 & 0.797 & 60.2\% & 0.016 & 0.061 & 0.942\\
         & F-MAML$_{dp}$\cite{Slack-FAT-2019} & 78.2\% & 0.025 & \textbf{0.114} & 0.943 & 97.6\% & 0.036 & 0.432 & 0.781 & 59.3\% & 0.013 & 0.058 & 0.945\\
         & F-MAML$_{eop}$\cite{Slack-FAT-2019} & 71.9\% & 0.028 & 0.134 & 0.927 & 94.1\% & 0.032 & 0.253 & 0.901 & 57.3\% & 0.012 & 0.054 & 0.941\\
         & LAFTR\cite{Madras-ICML-2018} & 72.3\% & 0.030 & 0.179 & 0.912 & 90.1\% & 0.050 & 0.401 & 0.790 & \textbf{62.3\%} & 0.025 & 0.098 & 0.877\\
         & Ours & \textbf{83.8\%} & \textbf{0.011} & 0.123 & \textbf{0.943} & 90.1\% & \textbf{0.016} & \textbf{0.215} & \textbf{0.927} & 61.3\% & 0.010 & \textbf{0.027} & \textbf{0.973}\\
         \hline
         
         \parbox[t]{2mm}{\multirow{8}{*}{\rotatebox[origin=c]{90}{\textbf{15-shot}}}} 
         & MAML\cite{Finn-ICML-2017-(MAML)} & \textbf{82.7\%} & 0.039 & 0.179 & 0.909 & \textbf{99.1\%} & 0.047 & 0.380 & 0.788 & 60.4\% & 0.016 & 0.068 & 0.903\\
         & Masked MAML & 80.2\% & - & 0.141 & 0.934 & 86.2\% & - & 0.246 & 0.870 & 58.1\% & - & 0.049 & 0.947 \\
         & pretrain & 80.6\% & 0.024 & 0.117 & 0.927 & 84.8\% & 0.029 & 0.264 & 0.859 & 57.6\% & 0.014 & 0.063 & 0.939\\
         & fair-MAML\cite{Zhao-ICKG-2-2020} & 65.4\% & 0.022 & 0.103 & 0.924 & 83.4\% & 0.021 & 0.221 & 0.895 & 56.2\% & 0.010 & 0.044 & 0.960\\
         & F-MAML$_{dp}$\cite{Slack-FAT-2019} & 81.0\% & 0.030 & 0.141 & 0.935 & 94.8\% & 0.039 & 0.313 & 0.812 & 57.9\% & 0.011 & 0.046 & 0.946\\
         & F-MAML$_{eop}$\cite{Slack-FAT-2019} & 80.8\% & 0.028 & 0.129 & 0.938 & 95.3\% & 0.040 & 0.320 & 0.815 & 58.4\% & 0.011 & 0.050 & 0.946 \\
         & LAFTR\cite{Madras-ICML-2018} & 75.5\% & 0.029 & 0.159 & 0.915 & 91.2\% & 0.030 & 0.299 & 0.825 & \textbf{61.1\%} & 0.012 & 0.089 & 0.892\\
         & Ours & 80.4\% & \textbf{0.005} & \textbf{0.011} & \textbf{0.985} & 80.6\% & \textbf{0.009} & \textbf{0.093} & \textbf{0.959} & 57.0\% & \textbf{0.005} & \textbf{0.010} & \textbf{0.989}\\
        \hline
         
        \parbox[t]{2mm}{\multirow{8}{*}{\rotatebox[origin=c]{90}{\textbf{20-shot}}}} 
         & MAML\cite{Finn-ICML-2017-(MAML)} & 82.5\% & 0.044 & 0.185 & 0.914 & \textbf{99.8\%} & 0.048 & 0.380 & 0.774 & 60.8\% & 0.014 & 0.062 & 0.912\\
         & Masked MAML & 80.8\% & - & 0.137 & 0.938 & 84.8\% & - & 0.242 & 0.876 & 57.8\% & - & 0.042 & 0.952\\
         & pretrain & 80.4\% & 0.021 & 0.100 & 0.935 & 84.9\% & 0.027 & 0.229 & 0.869 & 57.5\% & 0.012 & 0.053 & 0.942\\
         & fair-MAML\cite{Zhao-ICKG-2-2020} & 69.7\% & 0.018 & 0.083 & 0.931 & 86.0\% & 0.018 & 0.229 & 0.891 & 55.2\% & \textbf{0.005} & 0.044 & 0.964 \\
         & F-MAML$_{dp}$\cite{Slack-FAT-2019} & 80.6\% & 0.028 & 0.132 & 0.939 & 98.0\% & 0.042 & 0.314 & 0.816 & \textbf{67.4}\% & 0.010 & 0.042 & 0.951\\
         & F-MAML$_{eop}$\cite{Slack-FAT-2019} & \textbf{83.3\%} & 0.029 & 0.135 & 0.936 & 95.7\% & 0.038 & 0.318 & 0.817 & 58.1\% & 0.010 & 0.041 & 0.948\\
         & LAFTR\cite{Madras-ICML-2018} & 76.2\% & 0.032 & 0.175 & 0.911 & 89.8\% & 0.029 & 0.353 & 0.810 & 62.1\% & 0.015 & 0.095 & 0.875\\
         & Ours & 79.2\% & \textbf{0.001} & \textbf{0.018} & \textbf{0.988} & 85.7\% & \textbf{0.008} & \textbf{0.076} & \textbf{0.965} & 57.5\% & 0.006 & \textbf{0.006} & \textbf{0.991}\\
        \hline
    \end{tabular}
    \label{tab:result-cls}
\vspace{-5mm}
\end{table*}

\begin{enumerate}
    
    \item \textbf{MAML}: The model-agnostic meta-learning model with no fairness constraints proposed by \textit{Finn et al.,} \cite{Finn-ICML-2017-(MAML)}.
    
    \item \textbf{Masked MAML}: Similar to \textit{MAML}, this approach is applied to modified datasets by removing the protected attributes.
    
    \item \textbf{pretrain}: In computer vision, models pre-trained on large-scale image classification have been shown to learn effective features \cite{Donahue-ICML-2014}. In this paper, the pre-train baseline trains a single network on all tasks and in each task an unified fairness constraint is added to ensure DBC is satisfied.
    
    \item \textbf{fair-MAML}: \cite{Zhao-ICKG-2-2020} controls unfairness for each task and tunes a shared Lagrangian multiplier across tasks by simply applying grid search.
    
    \item \textbf{F-MAML$_{dp}$}: is a fair meta-learning approach proposed in \cite{Slack-FAT-2019}. In this baseline, \textit{Slack et al.,} proposed a simple regularization term aimed at achieving demographic parity for each task. All tasks share an unified regularization term in which the fairness hyperparameter is tuned through grid search, where the demographic parity regularizer $\mathcal{R}_{dp} = 1-p(\hat{y}=1|s=0)$.
    
    \item \textbf{F-MAML$_{eop}$}: is another fair meta-learning approach proposed in \cite{Slack-FAT-2019}, in which the demographic parity regularizer is replaced with the one aimed at improving equal opportunity, where $\mathcal{R}_{eop} = 1-p(\hat{y}=1|s=0, y=1)$.

    \item \textbf{LAFTR}\cite{Madras-ICML-2018}: is a transferring fair machine learning approach across domains that uses an adversarial approach to create an encoder that can be used to generate fair representations of datasets and demonstrate the utility of the encoder for fair transfer learning.
\end{enumerate}

\subsection{Experiment Setup and Parameter Tuning}
Our neural network trained follows the same architecture used by \cite{Finn-ICML-2017-(MAML)}, which contains 2 hidden layers of size of 40 with ReLU activation functions. When training, we use only one step gradient update (\textit{i.e.} $q=1$) and $k=10$ inner primal-dual updates with $2NK$ samples of query set, and a fixed primal and dual learning rate of $\gamma = 0.01$ and $\alpha = 0.01$. We use Adam as the meta-optimizer. Because we only consider a binary classification problem, all of tasks are 2-way, \textit{i.e. $N=2$}. Similarly, we set meta-learning rates of $\eta = 0.001$ and $\beta = 0.01$ used to update the meta-loss in the outer loop. For three datasets, all the unprotected attributes are standardized to zero mean and unit variance and prepared for experiments. Besides, taking few-shot learning into account, we set a meta batch-size of $8$ tasks and $4000$ meta-iterations for all datasets. Some key characteristics for all real data are listed in Table \ref{tab:key}.

All baseline models used to compare with our proposed approach share the same neural network architecture and parameter settings. Hyperparameters are selected by a held-out validation procedure. All experiments are repeated 10 times with the same settings. Results shown with these methods in this paper are mean of experimental outputs.

\section{Experiment Results}

\begin{figure*}[!htbp]
\captionsetup[subfigure]{aboveskip=-2pt,belowskip=-2pt}
\centering
    \begin{subfigure}[b]{0.325\textwidth}
        \includegraphics[width=\textwidth]{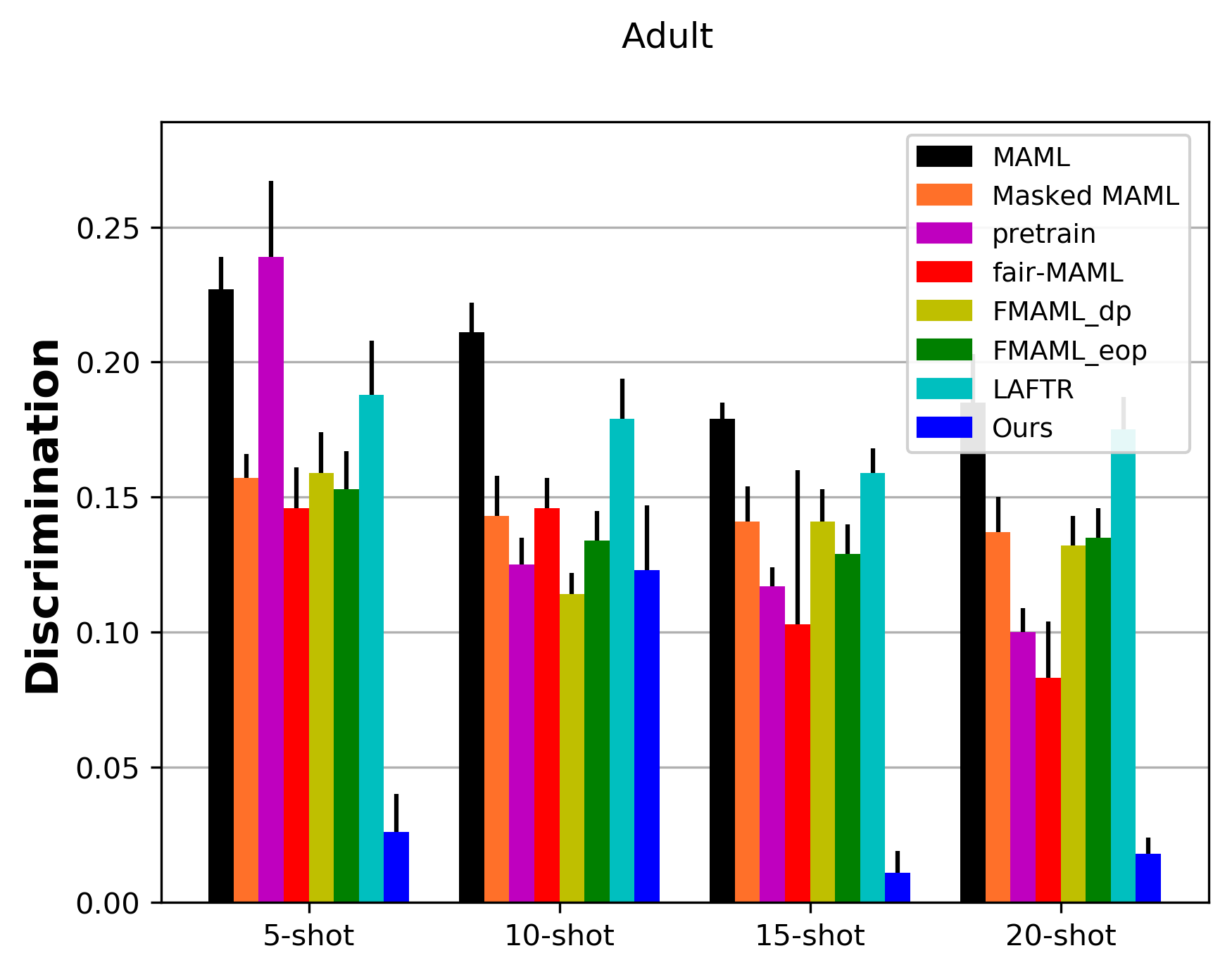}
        \caption{}
    \end{subfigure}
    \begin{subfigure}[b]{0.325\textwidth}
        \includegraphics[width=\textwidth]{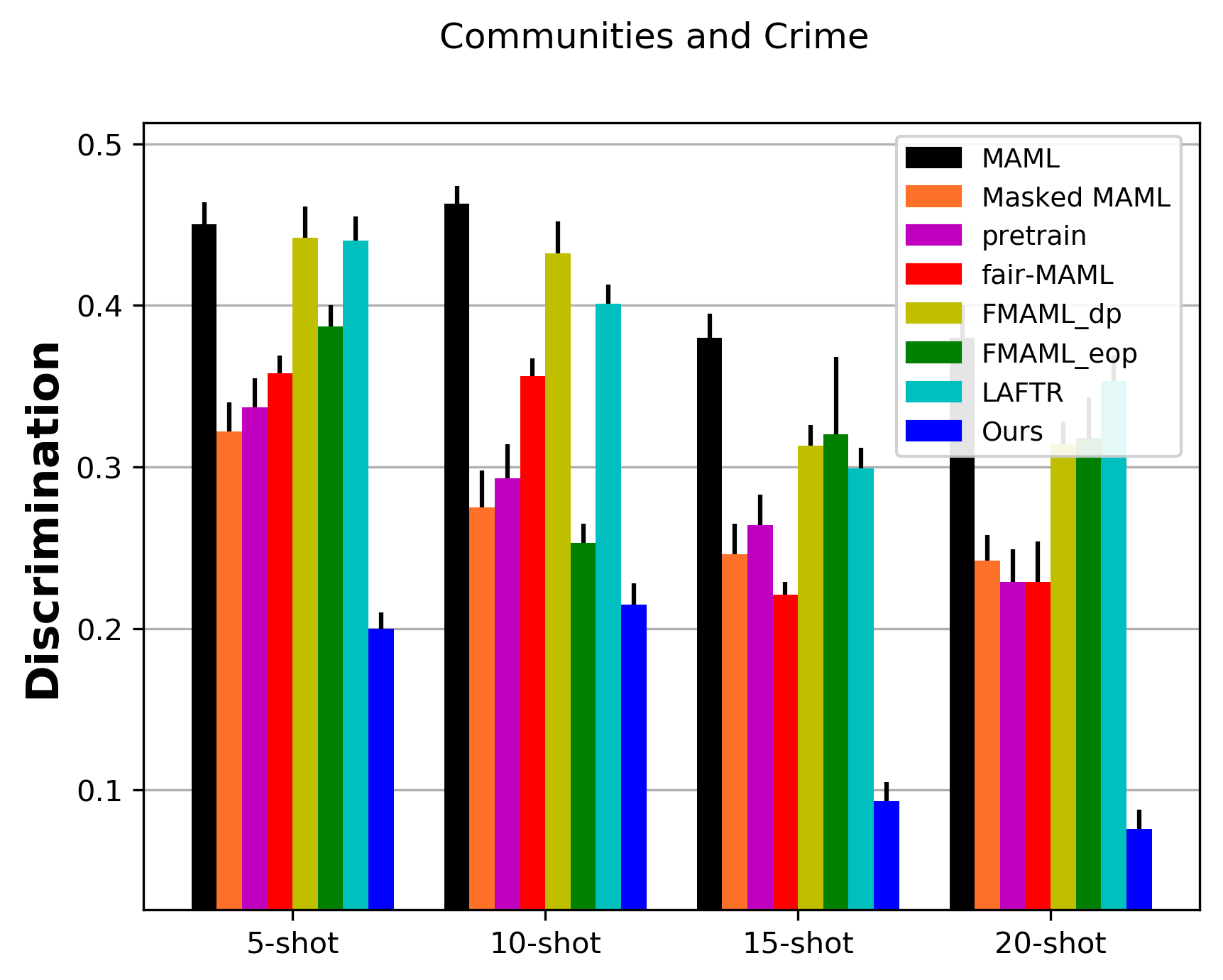}
        \caption{}
    \end{subfigure}
    \begin{subfigure}[b]{0.325\textwidth}
        \includegraphics[width=\textwidth]{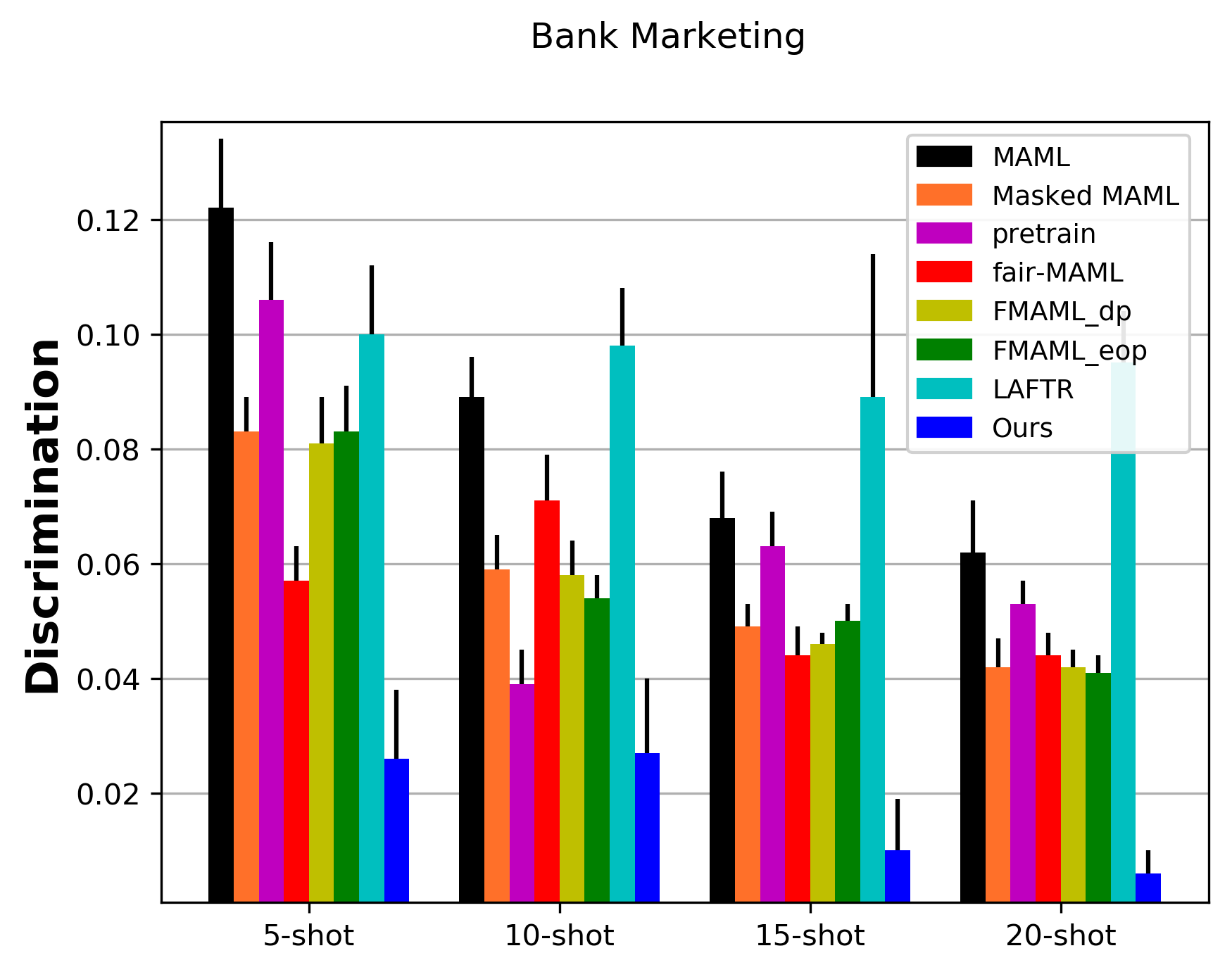}
        \caption{}
    \end{subfigure}
    
    \begin{subfigure}[b]{0.325\textwidth}
        \includegraphics[width=\textwidth]{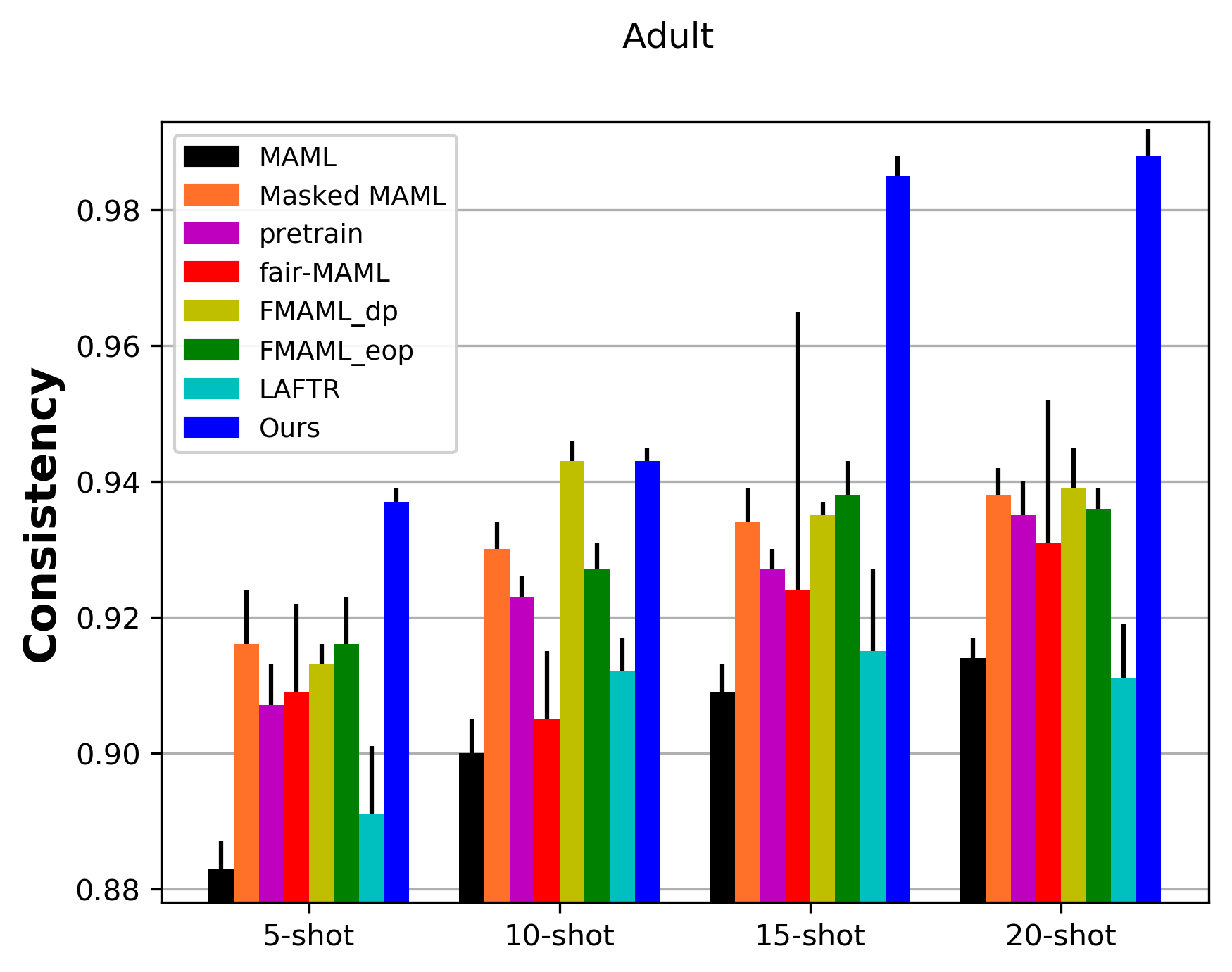}
        \caption{}
    \end{subfigure}
    \begin{subfigure}[b]{0.325\textwidth}
        \includegraphics[width=\textwidth]{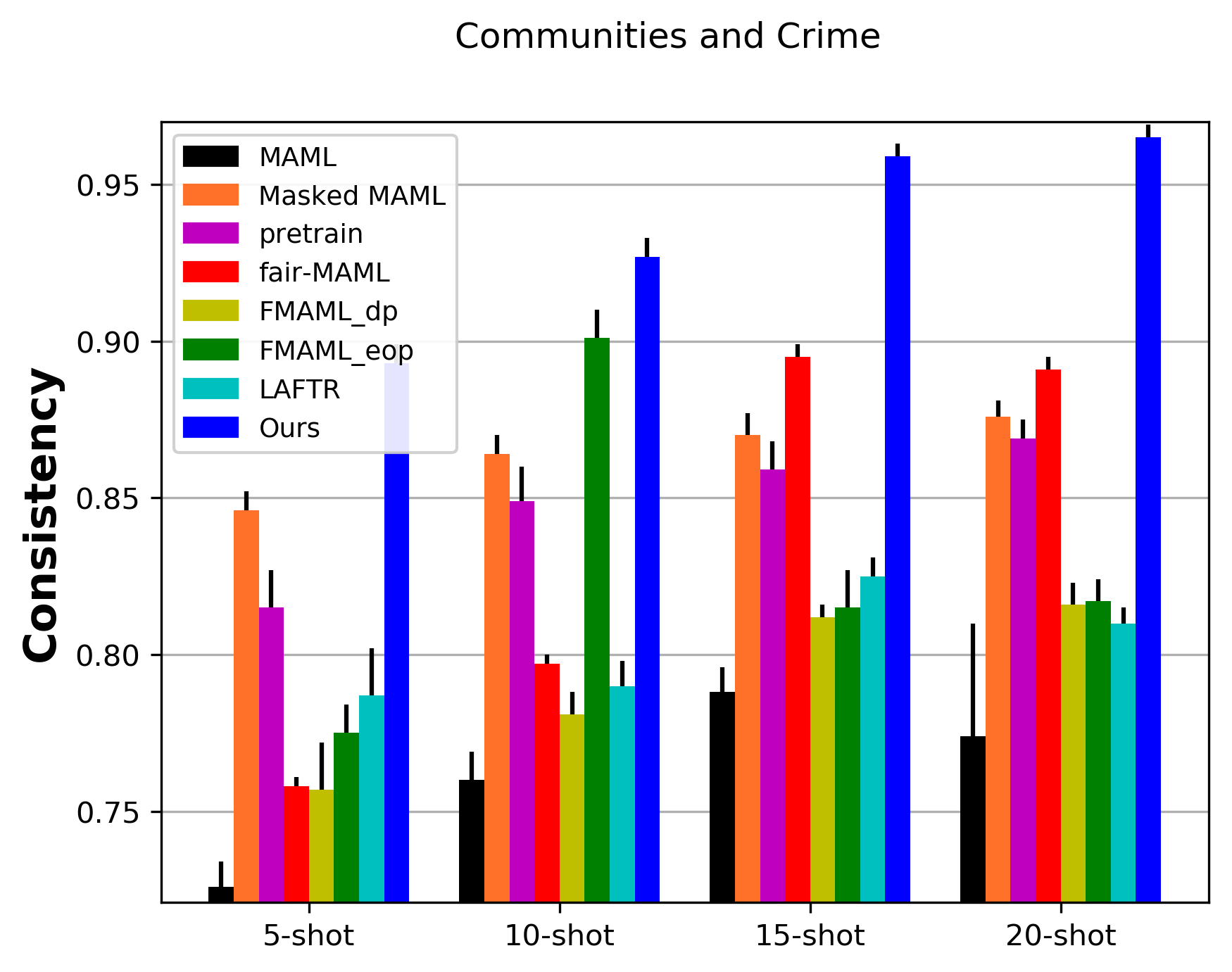}
        \caption{}
    \end{subfigure}
    \begin{subfigure}[b]{0.325\textwidth}
        \includegraphics[width=\textwidth]{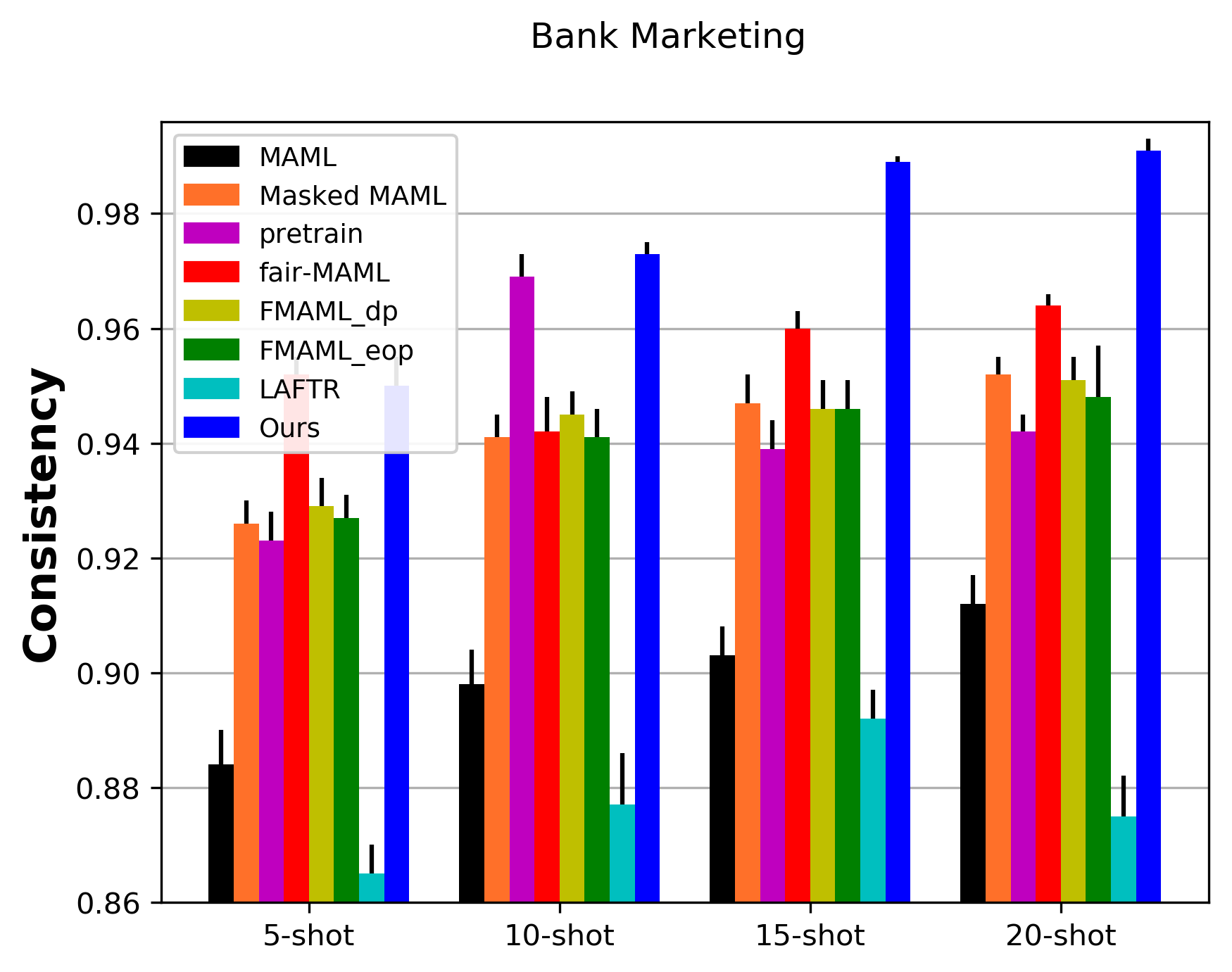}
        \caption{}
    \end{subfigure}
    
    \caption{Experiment results of real-world datasets in controlling biases.}
    \label{fig:evaluation metrics result}
\vspace{-5mm}
\end{figure*}

\begin{figure*}[!htbp]
\captionsetup[subfigure]{aboveskip=-2pt,belowskip=-2pt}
\centering
    \begin{subfigure}[b]{0.325\textwidth}
        \includegraphics[width=\textwidth]{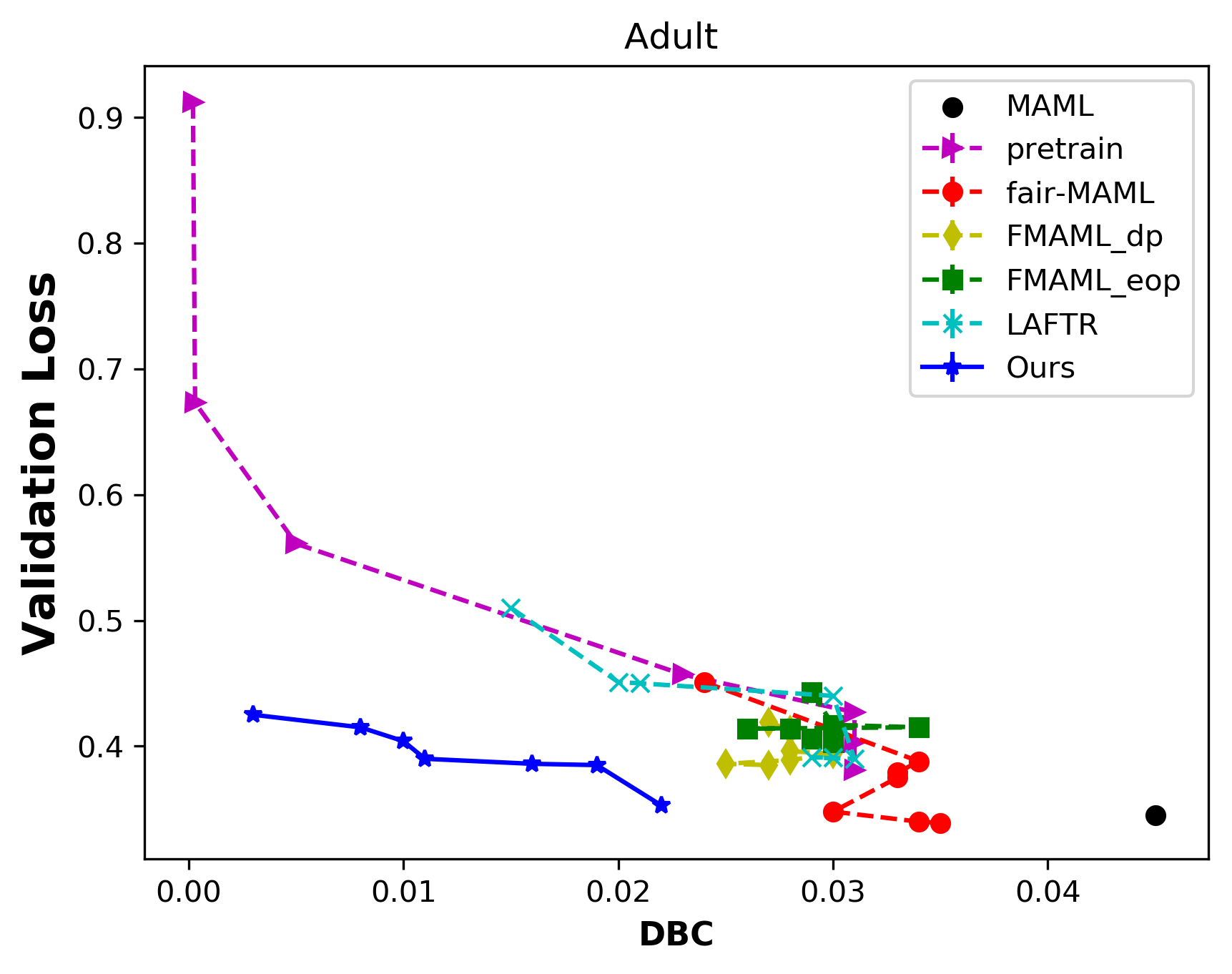}
        \caption{}
    \end{subfigure}
    \begin{subfigure}[b]{0.325\textwidth}
        \includegraphics[width=\textwidth]{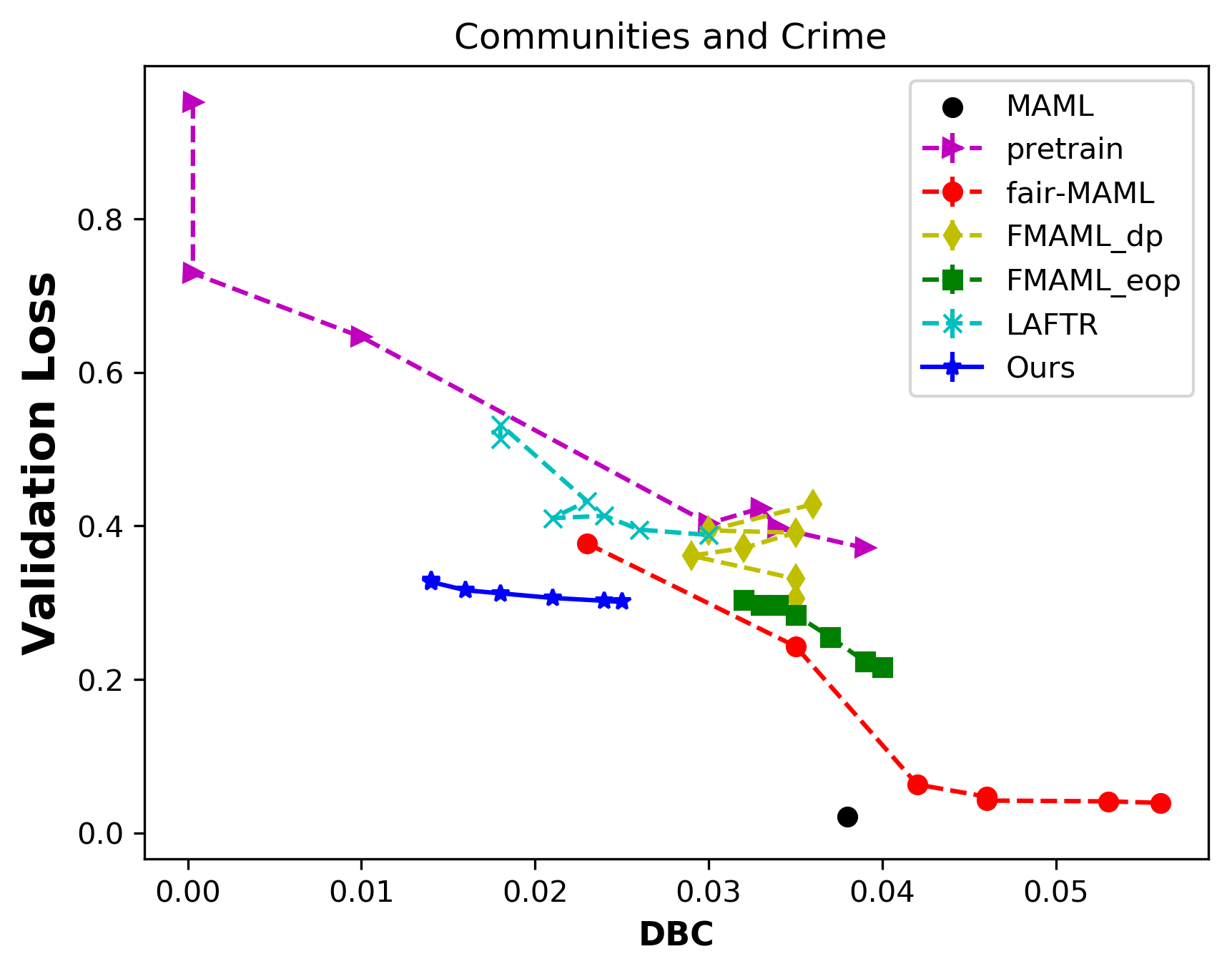}
        \caption{}
    \end{subfigure}
    \begin{subfigure}[b]{0.325\textwidth}
        \includegraphics[width=\textwidth]{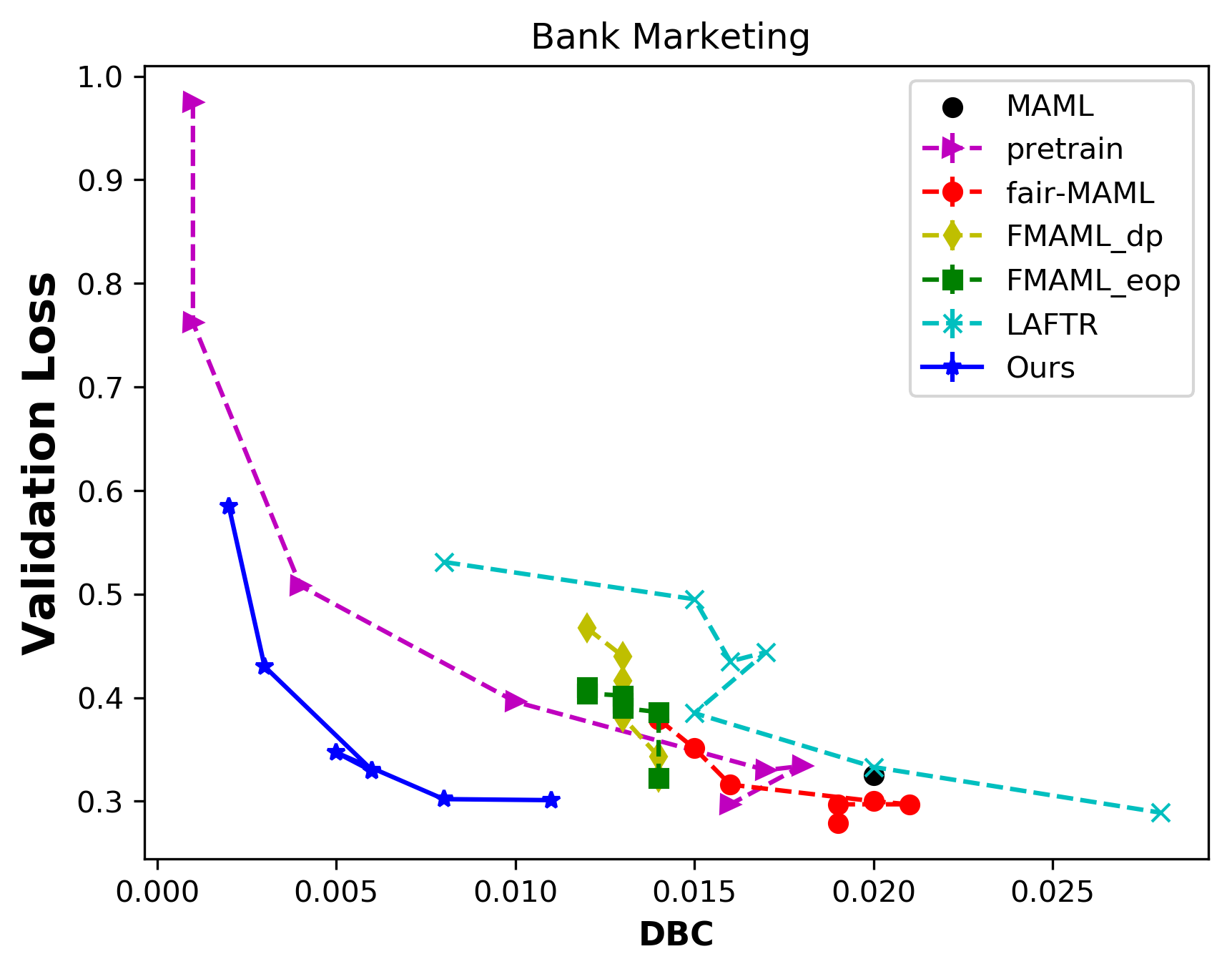}
        \caption{}
    \end{subfigure}
    \caption{The validation loss/fairness trade off sweeping over a range of dual variables.}
    \label{fig:trade off result}
\vspace{-5mm}
\end{figure*}

This section evaluates the effectiveness of the proposed approach and its competitors on a classification task. We focus on generalization of statistical parity on unseen tasks and trade-off between validation loss and fairness that the proposed dual subgradient method alleviates when used to train classifiers. For all baseline methods, wherever applicable, hyper-parameters were tuned via grid search. Specifically, we chose the models that were Pareto-optimal with regard to \textit{DBC} and all other evaluation metrics.

Consolidated and detailed performance of the different techniques over real-world data are listed in Table \ref{tab:result-cls}. We evaluate performance by fine-tuning the model learned by all methods on $K$-shot of $\{5, 10, 15, 20\}$ datapoints of each class for each dataset. Best performance in each experimental unit are labeled in bold. We first observe that there is a considerable amount of unfairness in the original datasets, which are reflected in the results of \textit{Data} in the table. Experiment results in Table \ref{tab:result-cls} demonstrates our proposed approach out-performs than other baseline methods in terms of controlling biases. It efficiently reduces \textit{DBC} from the original dataset and values of \textit{DBC} are limited to close zero that signify a fair prediction. In addition, fairness results based on two fair evaluation metrics, \textit{i.e.  Disc} (Figure \ref{fig:evaluation metrics result} (a-c)) and \textit{Cons} (Figure \ref{fig:evaluation metrics result} (d-f)), are plotted in Figure \ref{fig:evaluation metrics result}. Each trail was repeated 10 times and results shown in the figure are mean of experimental outputs followed by error bars representing one standard deviation of uncertainty.  

\textit{MAML} became a famous meta-learning algorithm because of its fast adaptation and good generalization performance on losses \cite{Finn-ICML-2017-(MAML)}. However, our results shows it fails to control biases nor performs success in fairness generalization in a few-shot meta-learning, although \textit{MAML} is stably able to produce high generalization accuracy. \textit{Masked MAML} shows an improvement in fairness; however, there is still substantial unfairness hidden in the data in the form of correlated attributes. \textit{F-MAML$_{dp}$} and \textit{F-MAML$_{eop}$} proposed by \textit{Slack et al.,} in \cite{Slack-FAT-2019} intuitively control unfairness by taking advantage of demographic parity and equal opportunity, respectively. Our results in Figure \ref{fig:evaluation metrics result} demonstrate that these two baseline methods fail to show fairness generalization onto unseen tasks in contrast to the proposed approach, in terms of reducing \textit{Disc} and promoting \textit{Cons}. Furthermore, though \textit{LAFTR} offers a way to transfer machine learning models between tasks, consistent with \cite{Slack-FAT-2019}, we observe it is unsuccessful in very data light situations. Besides, it is worth noting that we outperform baseline methods in bias controlling with better results as the number of training data increases. 

Although our proposed approach, \textit{PDFM}, returns a bit smaller predictive accuracies (see Table \ref{tab:result-cls}), this is due to the trade-off between losses and fairness. To this end, we train each method and sweep over a range of seven dual variables: $[0.001, 0.01, 0.1, 1, 10, 100, 1000]$. Taking 10-shot as an example, results presented in Figure \ref{fig:trade off result} is the mean across 10 runs on each set of dual variable using randomly selected hold out validation tasks. The fairness, \textit{i.e. DBC}, presented is the ratio between the protected and unprotected groups. Smaller validation loss and fairness values closer to zero (\textit{i.e.} bottom left in each sub-figure) indicate more successful outcomes. Here, as \textit{MAML} does not have hyper-parameters to control the loss/fairness trade-off, its outcomes across three datasets are presented with very low validation losses but high fairness values. In the proposed problem setting, the \textit{pretrain} neural network shows some ability to learn the new task using little data and fine-tuning epochs and as the dual variable increases, its validation losses decrease and thus \textit{DBC} increases. Moreover, \textit{LAFTR} is not successful at learning with minimal data and a small number of fine-tuning epochs for the new task. At low values,  \textit{fair-MAML, F-MAML$_{dp}$}, and \textit{F-MAML$_{eop}$} are able to achieve lower validation losses than the \textit{pretrain} and \textit{LAFTR} baselines. Crucially, the results stated in Figure \ref{fig:trade off result} confirm and further illustrate the findings that our proposed \textit{PDFM} is able to learn more accurate representations that are also fairer for the swept range than all baseline techniques.

\section{Conclusion and Future Work}
Techniques in meta-learning have been shown effectiveness for adaption of deep learning models on accuracy generalization to new tasks. These methods, however, are unable to ensure fairness adaption. In this paper, for the first time a novel Primal-Dual Fair Meta-learning (PDFM) framework is proposed, in which a good pair of primal-dual meta-parameters is optimally learned. To be specific, the meta-parameter pair is trained over a variety of learning tasks with a small amount of training samples. To produce the best performance, we implement two optimization strategies for both inner and meta subgradient update. Theoretical analysis justifies the efficiency and effectiveness of the proposed algorithms to support existence of solutions and algorithmic convergence guarantee. Results from extensive experiments demonstrate substantial improvements over the best prior work and our proposed framework is capable of generalization both accuracy and fairness onto new tasks. Further research in this area can make multitask parameters a standard ingredient in explainable fairness transfer learning.

\section*{Acknowledgement}
This work is supported by NSF awards IIS-1815696, IIS-1750911, DMS-1737978, DGE-2039542, and MRI-1828467.

\bibliographystyle{IEEEtran}
\bibliography{references}

\begin{thebibliography}{10}
\providecommand{\url}[1]{#1}
\csname url@samestyle\endcsname
\providecommand{\newblock}{\relax}
\providecommand{\bibinfo}[2]{#2}
\providecommand{\BIBentrySTDinterwordspacing}{\spaceskip=0pt\relax}
\providecommand{\BIBentryALTinterwordstretchfactor}{4}
\providecommand{\BIBentryALTinterwordspacing}{\spaceskip=\fontdimen2\font plus
\BIBentryALTinterwordstretchfactor\fontdimen3\font minus
  \fontdimen4\font\relax}
\providecommand{\BIBforeignlanguage}[2]{{%
\expandafter\ifx\csname l@#1\endcsname\relax
\typeout{** WARNING: IEEEtran.bst: No hyphenation pattern has been}%
\typeout{** loaded for the language `#1'. Using the pattern for}%
\typeout{** the default language instead.}%
\else
\language=\csname l@#1\endcsname
\fi
#2}}
\providecommand{\BIBdecl}{\relax}
\BIBdecl

\bibitem{Finn-ICML-2017-(MAML)}
C.~Finn, P.~Abbeel, and S.~Levine, ``Model-agnostic meta-learning for fast
  adaptation of deep networks,'' \emph{ICML}, 2017.

\bibitem{wang2020few}
Z.~Wang, Y.~Wang, Y.~Lin, E.~Delord, and K.~Latifur, ``Few-sample and
  adversarial representation learning for continual stream mining,'' in
  \emph{Proceedings of The Web Conference 2020}, 2020, pp. 718--728.

\bibitem{Zafar-AISTATS-2017}
M.~B. Zafar, I.~Valera, M.~G. Rodriguez, and K.~P. Gummadi, ``Fairness
  constraints: Mechanisms for fair classification,'' \emph{AISTATS}, 2017.

\bibitem{yoon2018}
J.~Yoon, T.~Kim, O.~Dia, S.~Kim, Y.~Bengio, and S.~Ahn, ``Bayesian
  model-agnostic meta-learning,'' in \emph{NeurIPS}, 2018, pp. 7332--7342.

\bibitem{Vinyals-NIPS-2016-(MatchingNet)}
O.~Vinyals, C.~Blundell, T.~Lillicrap, K.~Kavukcuoglu, and D.~Wierstra,
  ``Matching networks for one shot learning,'' \emph{NeurIPS}, 2016.

\bibitem{Snell-NIPS-2017-(ProtoNet)}
J.~Snell, K.~Swersky, and R.~Zemel, ``Prototypical networks for few-shot
  learning,'' \emph{NeurIPS}, 2017.

\bibitem{xu2018meta}
Z.~Xu, H.~P. van Hasselt, and D.~Silver, ``Meta-gradient reinforcement
  learning,'' in \emph{NeurIPS}, 2018, pp. 2396--2407.

\bibitem{Ravi-ICLR-2017}
S.~Ravi and H.~Larochelle, ``Optimization as a model for few-shot learning,''
  \emph{ICLR}, 2017.

\bibitem{Finn-NIPS-2018}
C.~Finn, K.~Xu, and S.~Levine, ``Probabilistic model-agnostic meta-learning,''
  in \emph{NeurIPS}, 2018, pp. 9516--9527.

\bibitem{Finn-ICML-2019}
C.~Finn, A.~Rajeswaran, S.~Kakade, and S.~Levine, ``Online meta-learning.''
  \emph{ICML}, 2019.

\bibitem{Nichol-arXiv-Reptile-2018}
A.~Nichol and J.~Schulman, ``Reptile: a scalable metalearning algorithm,''
  \emph{arXiv preprint arXiv:1803.02999}, 2018.

\bibitem{Rusu-ICLR-2019}
A.~A. Rusu, D.~Rao, J.~Sygnowski, O.~Vinyals, R.~Pascanu, S.~Osindero, and
  R.~Hadsell, ``Meta-learning with latent embedding optimization,''
  \emph{ICLR}, 2019.

\bibitem{Zliobaite-arXiv-2015}
I.~Zliobaite, ``A survey on measuring indirect discrimination in machine
  learning.'' \emph{arXiv preprint arXiv:1511.00148}, 2015.

\bibitem{Zemel-ICML-2013}
R.~Zemel, Y.~Wu, K.~Swersky, T.~Pitassi, and C.~Dwork, ``Learning fair
  representations,'' \emph{ICML}, 2013.

\bibitem{Palomar-dual_decomposition-2007}
D.~P. Palomar and M.~Chiang, ``Alternative distributed algorithms for network
  utility maximization: framework and applications.'' \emph{IEEE Transactions
  on Automatic Control}, vol. 52(12), pp. 2254--2269, 2007.

\bibitem{Raffard-dual_decomposition-2004}
R.~L. Raffard, C.~J. Tomlin, , and S.~P. Boyd, ``Distributed optimization for
  cooperative agents: Application to formation flight.'' \emph{In Proceedings
  of the IEEE Conference on Decision and Control}, pp. 2453--2459, 2004.

\bibitem{Bengio-ICLR-2020}
Y.~Bengio, T.~Deleu, N.~Rahaman, N.~R. Ke, S.~Lachapelle, O.~Bilaniuk,
  A.~Goyal, and C.~Pa, ``A meta-transfer objective for learning to disentangle
  causal mechanisms,'' \emph{ICLR}, 2020.

\bibitem{Yao-ICML-2019}
H.~Yao, Y.~Wei, J.~Huang, and Z.~Li, ``Hierarchically structured
  meta-learning,'' \emph{ICML}, 2019.

\bibitem{Tseng-ICLR-2020}
H.-Y. Tseng, H.-Y. Lee, J.-B. Huang, and M.-H. Yang, ``Cross-domain few-shot
  classification via learned feature-wise transformation,'' \emph{ICLR}, 2020.

\bibitem{Lian-ICLR-2020}
D.~Lian, Y.~Zheng, Y.~Xu, Y.~Lu, L.~Lin, P.~Zhao, J.~Huang, and S.~Gao,
  ``Towards fast adaptation of neural architectures with meta learning,''
  \emph{ICLR}, 2020.

\bibitem{sung2018learning}
F.~Sung, Y.~Yang, L.~Zhang, T.~Xiang, P.~H. Torr, and T.~M. Hospedales,
  ``Learning to compare: Relation network for few-shot learning,'' in
  \emph{CVPR}, 2018, pp. 1199--1208.

\bibitem{wang2019robust}
Z.~Wang, Z.~Kong, S.~Changra, H.~Tao, and L.~Khan, ``Robust high dimensional
  stream classification with novel class detection,'' in \emph{ICDE}, 2019, pp.
  1418--1429.

\bibitem{Antoniou-ICLR-2019}
A.~Antoniou, H.~Edwards, and A.~Storkey, ``How to train your maml,''
  \emph{ICLR}, 2019.

\bibitem{franceschi2018bilevel}
L.~Franceschi, P.~Frasconi, S.~Salzo, R.~Grazzi, and M.~Pontil, ``Bilevel
  programming for hyperparameter optimization and meta-learning,'' \emph{ICML},
  2018.

\bibitem{Calders-ICDM-2013}
T.~Calders, A.~Karim, F.~Kamiran, W.~Ali, and X.~Zhang, ``Controlling attribute
  effect in linear regression,'' \emph{ICDM}, 2013.

\bibitem{Feldman-KDD-2015}
M.~Feldman, S.~Friedler, J.~Moeller, C.~Scheidegger, and S.~Venkatasubramanian,
  ``Certifying and removing disparate impact.'' \emph{KDD}, 2015.

\bibitem{Hardt-NIPS-2016}
M.~Hardt, E.~Price, and N.~Srebro, ``Equality of opportunity in supervised
  learning.'' \emph{NeurIPS}, 2016.

\bibitem{Berk-FATML-2018}
R.~Berk, H.~Heidari, S.~Jabbari, M.~Joseph, M.~Kearns, J.~Morgenstern, S.~Neel,
  and A.~Roth, ``A convex framework for fair regression.'' \emph{FAT ML}, 2018.

\bibitem{Zhao-ICDM-2019}
C.~Zhao and F.~Chen, ``Rank-based multi-task learning for fair regression,''
  \emph{IEEE International Conference on Data Mining (ICDM)}, 2019.

\bibitem{Gondek-KDD-2005}
D.~Gondek and T.~Hofman, ``Non-redundant clustering with conditional
  ensembles.'' \emph{KDD}, 2005.

\bibitem{Kamishima-RR-w-2017}
T.~Kamishima and S.~Akaho, ``Considerations on recommendation independence for
  a find-good-items task.'' \emph{In Workshop on Responsible Recommendation},
  2017.

\bibitem{singh2018}
A.~Singh and T.~Joachims, ``Fairness of exposure in rankings,'' in
  \emph{KDD(2018)}, pp. 2219--2228.

\bibitem{Slack-FAT-2019}
D.~Slack, S.~Friedler, and E.~Givental, ``Fairness warnings and fair-maml:
  Learning fairly with minimal data,'' \emph{Proceedings of the Conference on
  Fairness, Accountability, and Transparency (FAT)}, 2020.

\bibitem{Zhao-ICKG-1-2020}
C.~Zhao and F.~Chen, ``Unfairness discovery and prevention for few-shot
  regression.'' \emph{ICKG}, 2020.

\bibitem{Zhao-ICKG-2-2020}
C.~Zhao, C.~Li, J.~Li, and F.~Chen, ``Fair meta-learning for few-shot
  classification.'' \emph{ICKG}, 2020.

\bibitem{Zhang-AAAI-2019}
L.~Zhang, Y.~Wu, and X.~Wu, ``Fairness-aware classification: Criterion,
  convexity, and bounds.'' \emph{AAAI}, 2019.

\bibitem{Goel-AAAI-2018}
N.~Goel, M.~Yaghini, , and B.~Faltings, ``Non-discriminatory machine learning
  through convex fairness criteria.'' \emph{AAAI}, 2018.

\bibitem{Rush-AIR-2012}
A.~Rush and M.~Collins, ``A tutorial on dual decomposition and lagrangian
  relaxation for inference in natural language processing.'' \emph{Journal of
  Artificial Intelligence Research}, 2012.

\bibitem{Bertsekas-book}
D.~P. Bertsekas, ``Nonlinear programming,'' \emph{Journal of the Operational
  Research Society}, vol.~48, no.~3, pp. 334--334, 1997.

\bibitem{averagedDS-SJO-2009}
A.~Nedic and A.~Ozdaglar, ``Approximate primal solutions and rate analysis for
  dual subgradient methods,'' \emph{SIAM Journal on Optimization}, vol.~19, pp.
  1757--1780, 01 2009.

\bibitem{Fallah-AISTATS-2020}
A.~Fallah, A.~Mokhtari, and A.~Ozdaglar, ``On the convergence theory of
  gradient-based model-agnostic meta-learning algorithms.'' \emph{AISTATS},
  2020.

\bibitem{AdultDataSet-UCI-1994}
R.~Kohavi and B.~Becker, ``Uci machine learning repository,'' 1994.

\bibitem{CommunitiesandCrimeDataSet-UCI-1994}
M.~Lichman, ``Uci machine learning repository,'' 2013.

\bibitem{Moro-bank-marketing-dataset-2014}
S.~Moro, P.~Cortez, and P.~Rita, ``A data-driven approach to predict the
  success of bank telemarketing.'' \emph{Decision Support Systems}, 2014.

\bibitem{Madras-ICML-2018}
D.~Madras, E.~Creager, T.~Pitassi, and R.~Zemel, ``Learning adversarially fair
  and transferable representations.'' \emph{ICML}, 2018.

\bibitem{Donahue-ICML-2014}
J.~Donahue, Y.~Jia, O.~Vinyals, J.~Hoffman, N.~Zhang, E.~Tzeng, and T.~Darrell,
  ``Decaf: A deep convolutional activation feature for generic visual
  recognition.'' \emph{ICML}, 2014.

\end{thebibliography}

\end{document}